\documentclass[sigconf,nonacm]{acmart}

\AtBeginDocument{%
  }

\setcopyright{acmlicensed} 
\copyrightyear{2026}
\acmYear{2026}
\acmDOI{XXXXXXX.XXXXXXX} 

\acmConference[KDD '26]{Proceedings of the 32nd ACM SIGKDD Conference on Knowledge Discovery and Data Mining}
{August 9--13, 2026}{Jeju, Republic of Korea}

\acmISBN{978-1-4503-XXXX-X/2026/08} 

\renewcommand\footnotetextcopyrightpermission[1]{} 

\usepackage{tcolorbox}
\tcbuselibrary{skins, breakable}
\usepackage{booktabs}
\usepackage{longtable}
\usepackage{array}

\tcbuselibrary{skins} 

\newtcolorbox{promptbox}[1][]{%
  enhanced,                 
  colback=gray!5,           
  colframe=gray!50,         
  coltitle=black,           
  fonttitle=\bfseries,      
  title={#1},               
  attach boxed title to top left={xshift=2mm, yshift=-2mm}, 
  sharp corners=downhill,   
  arc=0.5mm,                
  boxrule=0.5pt,            
  left=2mm, right=2mm, top=2mm, bottom=2mm, 
  fontupper=\footnotesize\ttfamily, 
}
\usepackage{algorithm}
\usepackage{algpseudocode}
\usepackage{amsmath} 
\usepackage{multirow}
\usepackage{xcolor}
\usepackage[table]{xcolor}

\newcommand{\rise}[1]{\textcolor{green!65!black}{\scriptsize{(+#1)}}}
\definecolor{mygreen}{HTML}{056b34}
\definecolor{myblue}{HTML}{054488}
\definecolor{myyellow}{HTML}{C9A227}
\definecolor{myred}{HTML}{8B1E3F}
\newcommand{\green}[1]{%
  \begingroup
  \setlength{\fboxsep}{1pt}
  \setlength{\fboxrule}{0.8pt}
  \fcolorbox{mygreen}{white}{\textcolor{mygreen}{#1}}%
  \endgroup
}
\newcommand{\blue}[1]{%
  \begingroup
  \setlength{\fboxsep}{1pt}
  \setlength{\fboxrule}{0.8pt}
  \fcolorbox{myblue}{white}{\textcolor{myblue}{#1}}%
  \endgroup
}
\newcommand{\yellow}[1]{%
  \begingroup
  \setlength{\fboxsep}{1pt}
  \setlength{\fboxrule}{0.8pt}
  \fcolorbox{myyellow}{white}{\textcolor{myyellow}{#1}}%
  \endgroup
}
\newcommand{\red}[1]{%
  \begingroup
  \setlength{\fboxsep}{1pt}
  \setlength{\fboxrule}{0.8pt}
  \fcolorbox{myred}{white}{\textcolor{myred}{#1}}%
  \endgroup
}
\begin{document}

\title{DeepRead: Document Structure-Aware Reasoning to Enhance Agentic Search}

\author{
  Zhanli Li$^{3*}$,
  Huiwen Tian$^{1,2}$,
  Lvzhou Luo$^{1,2}$,
  Yixuan Cao$^{1,2\dagger}$,
  Ping Luo$^{1,2}$
}

\affiliation{%
  \institution{$^{1}$ State Key Laboratory of AI Safety, Institute of Computing Technology,
  Chinese Academy of Sciences (CAS)}
  \city{Beijing}
  \postcode{100190}
  \country{China}
}

\affiliation{%
  \institution{$^{2}$ University of Chinese Academy of Sciences, CAS}
  \city{Beijing}
  \postcode{100049}
  \country{China}
}

\affiliation{%
  \institution{$^{3}$ Wenlan School of Business, Zhongnan University of Economics and Law}
  \city{Wuhan}
  \postcode{430073}
  \country{China}
}

\email{lizhanli@stu.zuel.edu.cn, tianhuiwen25@mails.ucas.ac.cn, {luolvzhou23s, caoyixuan, luop}@ict.ac.cn}
\renewcommand{\shortauthors}{Li et al.}


\begin{abstract}
With the rapid advancement of tool-use capabilities in Large Language Models (LLMs), Retrieval-Augmented Generation (RAG) is shifting from static, one-shot retrieval toward autonomous, multi-turn evidence acquisition. However, existing agentic search frameworks typically treat long documents as flat collections of unstructured chunks, disregarding the native hierarchical organization and sequential logic essential for human comprehension. To bridge this gap, we introduce \textbf{DeepRead}, a structure-aware document reasoning agent designed to operationalize document-native structural priors into actionable reasoning capabilities. Leveraging the structural fidelity of modern OCR, DeepRead constructs a paragraph-level, coordinate-based navigation system and equips the LLM with two synergistic tools: \textsf{Retrieve} for scanning-aware localization, and \textsf{ReadSection} for contiguous, order-preserving reading within specific hierarchical scopes. This design elicits a human-like ``locate-then-read'' reasoning paradigm, effectively mitigating the context fragmentation inherent in traditional retrieval methods. Extensive evaluations across four benchmarks spanning diverse document types demonstrate that DeepRead outperforms Search-o1-style agentic search baselines by an average of 10.3\%. Fine-grained behavioral analysis further confirms that DeepRead autonomously adopts human-aligned reading strategies, validating the critical role of structural awareness in achieving precise document reasoning. Our code is available at \url{https://github.com/Zhanli-Li/DeepRead}.
\end{abstract}




\maketitle

\begingroup
\renewcommand\thefootnote{\fnsymbol{footnote}}
\setcounter{footnote}{0}
\footnotetext[1]{The first author contributed to this work during an internship at State Key Laboratory of AI Safety, Institute of Computing Technology,
Chinese Academy of Sciences (CAS). The paper is currently in preview form, and this work is still in progress.} 
\footnotetext[2]{Corresponding author.} 
\endgroup
\section{Introduction}
\begin{figure}[t]
    \centering
    \includegraphics[width=\columnwidth]{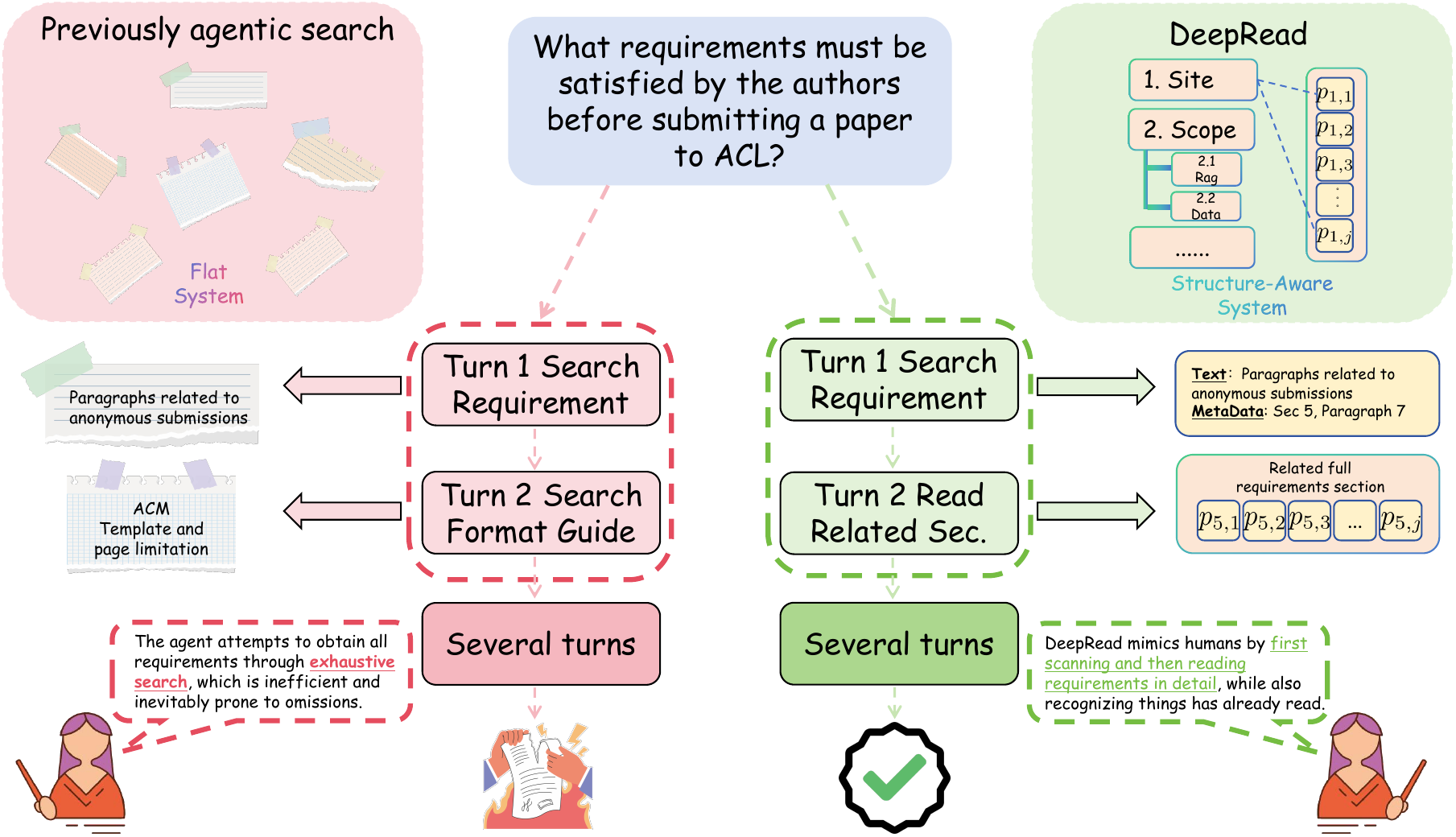}
    \caption{A Comparison of Search-o1-style Agentic Search and DeepRead on a Toy Case}
    \label{fig:toy-intro}
\end{figure}

When humans seek knowledge from documents, they rarely rely on a single, linear scan or random keyword matching. Instead, they employ a structured ``locate-then-read'' strategy: first, roughly locate the position, then proceed with close reading. In contrast, while LLMs have achieved impressive performance in general natural language understanding, they remain brittle when attempting to replicate this precise, evidence-based reasoning — a limitation that has spurred the development of retrieval-augmented methods to enhance reliability. Two factors are particularly constraining: (i) static parametric memory cannot faithfully encode ever-changing or domain-specific details, and (ii) LLMs tend to produce plausible but unsupported statements (hallucinations) when evidence is missing. RAG mitigates these issues by grounding generation in external sources~\cite{lewis2020retrieval}, yet standard approaches often fail to capture the structural reading priors that humans naturally possess—a gap that subsequent research has sought to address through evolving RAG frameworks.

Early RAG systems predominantly adopted one-shot pipelines, where retrieval is executed once and the answer is generated from a fixed set of top-ranked chunks. Research in this phase focused on improving retrieval precision through stronger embedding models, optimized indexing, and coarse-to-fine reranking. While these advances improved single-step retrieval accuracy, the interaction pattern remained static: the system neither revises its information needs nor adapts its access strategy as reasoning unfolds. This limitation becomes pronounced for long-document and multi-hop scenarios, where evidence is widely distributed and cannot be reliably captured by a single retrieval call. Recent work such as \citet{zhao2024longrag} mitigates the ``lost-in-the-middle'' phenomenon for long contexts, but it largely remains within fixed, single-round retrieval pipelines and lacks the interactivity required for complex reasoning. 

To better handle multi-step dependencies, approaches such as PlanRAG~\cite{lee2024planrag} introduced explicit planning followed by retrieval. However, such two-stage designs can be brittle: they depend heavily on the quality of the initial plan and have limited ability to adapt when intermediate findings deviate from expectations. In parallel, iterative retrieval methods~\cite{feng2024retrieval} have emerged to gather information across multiple turns. Nevertheless, many of these methods still follow \emph{prescribed} schedules (e.g., a fixed number of rounds or a rigid retrieve--read--generate loop). Such rigidity can be inefficient for simple queries that require minimal evidence, and insufficient for complex queries that demand extensive, adaptive evidence acquisition~\cite{feng2024retrieval}. More recently, as major foundation-model providers have increasingly emphasized \emph{agentic} capabilities---especially the ability to invoke external tools flexibly and accurately---\emph{agentic RAG} has reshaped this landscape by casting evidence acquisition as an autonomous decision-making process driven by tool use. In frameworks such as Search-o1~\cite{li2025search}, the LLM is no longer bound to a fixed retrieve--generate schedule; instead, it can decide \emph{when to search}, \emph{what to search for}, and \emph{when sufficient evidence has been gathered to stop and answer}. This autonomy enables markedly more adaptive information seeking than classical iterative retrieval, allowing the model to adjust its trajectory on the fly based on intermediate reasoning signals and feedback from retrieved results.

Despite this progress, as illustrated in the toy example of Fig.~\ref{fig:toy-intro}, Search-o1-style agentic search remains \emph{structurally blind} when operating over long and organized documents. Consider the query: ``What requirements must be satisfied by the authors before submitting a paper to ACL?'' A structure-agnostic agent is forced into a cycle of \emph{keyword exhaustion and guesswork}: it repeatedly issues narrowly specified searches (e.g., `format', `anonymity', `template') and stitches together disjoint snippets. This strategy is inherently omission-prone—if the agent fails to hypothesize a particular keyword (e.g., `page limit'), that requirement is simply missed—and, without a notion of \emph{examined regions}, it often wastes turns redundantly revisiting content it has effectively already covered. Yet for long-context information needs, evidence is rarely scattered randomly; it is typically organized systematically within a dedicated region (e.g., a ``Submission Guideline'' section). Intuitively, once the agent encounters \emph{any} indicative clue that lands it inside the correct section (such as a sentence describing the anonymity policy), the optimal behavior should shift from further keyword guessing to \emph{contiguous, order-preserving reading} of the surrounding section. In other words, a single localized snippet should act as a structural anchor that triggers sequential reading over the co-located neighborhood, enabling comprehensive capture of all nearby requirements—including those never explicitly queried—in one pass, while avoiding the retrieval noise and context fragmentation that often arise in preprocessing-heavy pipelines~\cite{gong2025mmrag}.

Meanwhile, recent open-source OCR models have achieved remarkable success, enabling accurate recovery of document hierarchy and reading order from visually rich inputs~\cite{cui2025paddleocr,wei2025deepseek,team2025hunyuanocr,wang2024mineru,li2025dots}. Trained on large-scale document images paired with structured markup (e.g., Markdown), these models can extract paragraph-level organization, where Markdown naturally preserves both hierarchical layout (e.g., headings and lists) and sequential flow—structural priors intentionally encoded by authors and central to human document comprehension. This progress makes document-native priors increasingly accessible to downstream LLM reasoning systems. However, most existing agentic search frameworks still interact with documents through structure-agnostic chunk collections, which offers little support for hierarchy- or order-aware comprehension. As a result, even when agentic search enables flexible ``search-and-think'' behavior, the underlying interface remains largely blind to document topology—especially hierarchical and sequential cues—limiting robust long-document reasoning.

Leveraging the success of open-source OCR models and the significant enhancement of LLM agentic capabilities, we propose \textbf{DeepRead}, a structure-aware document reasoning agent that operationalizes document hierarchy and sequential priors for multi-turn QA. DeepRead builds upon the autonomous decision-making paradigm of agentic search but fixes a key bottleneck---the lack of document-native topology in the interaction interface. Specifically, DeepRead maps each document into a \emph{structural coordinate system} (section and paragraph indices) and equips the LLM with two synergistic tools: \textsf{Retrieve}, which performs scanning-aware localization and returns coordinate-anchored evidence, and \textsf{ReadSection}, which enables contiguous, order-preserving reading within a specified section and paragraph range. This interface supports a human-like ``\emph{locate-then-read}'' pattern: first pinpointing relevant regions via lightweight scanning, then consuming complete local narratives. This design improves navigation efficiency over long, structured texts and mitigates the context fragmentation inherent in flat retrieval paradigms. Our contributions are summarized as follows:
\begin{itemize}
    \item  We propose a coordinate-based reasoning framework that operationalizes document hierarchy and sequence. By coupling \textsf{Retrieve} with \textsf{ReadSection}, we enable an emergent \emph{locate-then-read} paradigm that reconstructs contiguous evidence from fragmented search results.
    
    \item Extensive evaluations across four benchmarks—spanning financial analysis and multi-document reasoning—show that DeepRead outperforms Search-o1-style baselines by an average of \textbf{10.3\%} accuracy, demonstrating exceptional efficacy in handling long-range dependencies and cross-document integration.
    
    \item Through fine-grained analysis, we validate that DeepRead exhibits human-aligned reading patterns, balancing targeted search with sequential reading. Ablation studies further confirm the critical synergy between \textsf{Retrieve} and \textsf{Read} tools, particularly in complex multi-document scenarios.
\end{itemize}

\begin{figure*}[t]
    \centering
    \includegraphics[width=1\textwidth]{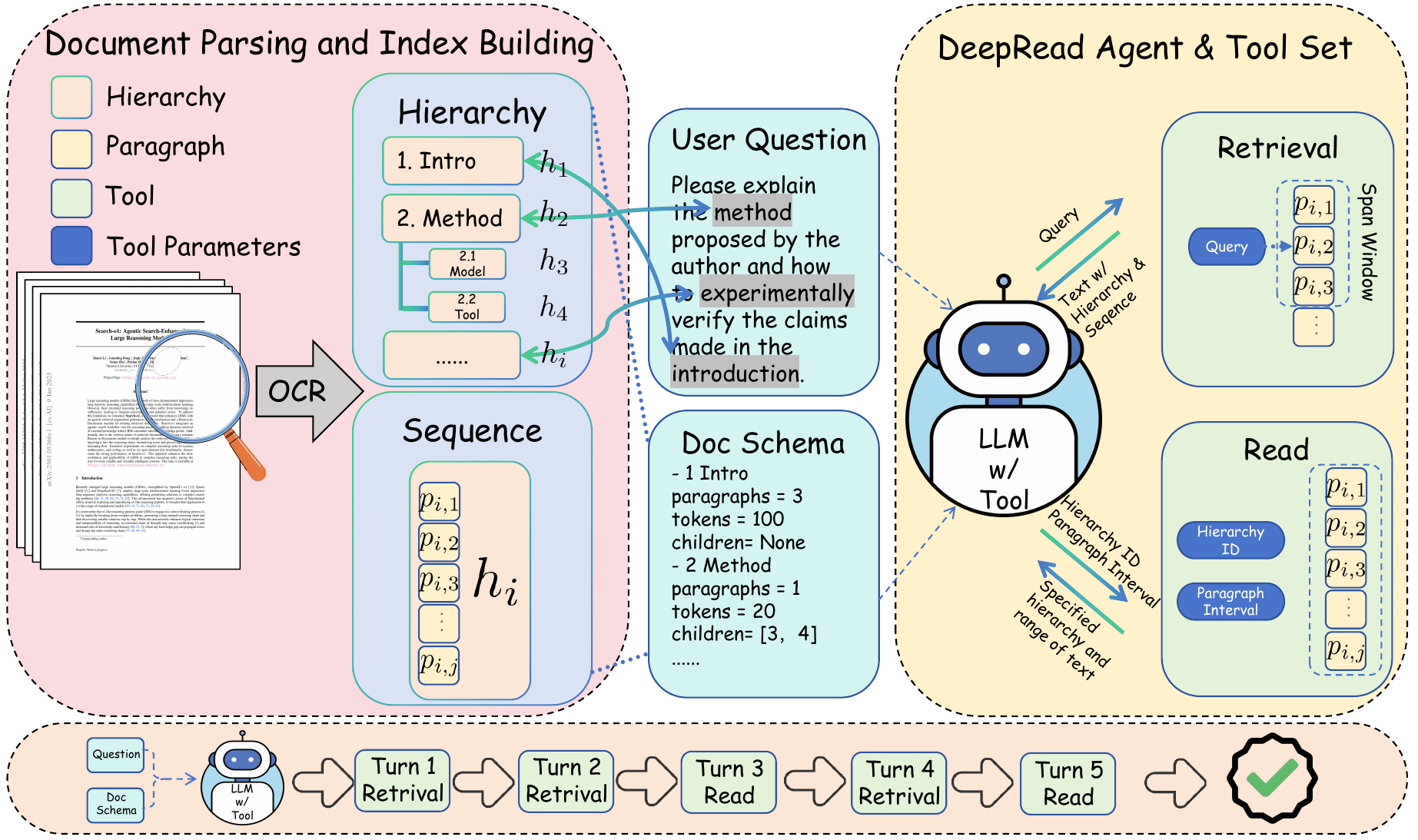}
    \caption{This is the DeepRead framework diagram. It takes user questions parsed into Doc Schema as input. Using two tools, LLM can perform multi-turn tool invocations to answer user questions.}
    \label{fig:main-deepread}
\end{figure*}

\section{Related Work}
\textbf{Document QA and RAG.} Document Question Answering (DocQA) is generally divided into \emph{open} and \emph{closed} settings. While open QA retrieves evidence from massive external corpora, closed QA restricts the scope to specific documents, prioritizing \emph{precise localization} and \emph{faithful interpretation}—critical for legal, financial, and scientific domains. To tackle the challenge of locating evidence in long contexts, Retrieval-Augmented Generation (RAG) has become the standard solution. Early \emph{Naive RAG} systems relied on single-pass top-$k$ retrieval, which often falls short in complex reasoning tasks. Consequently, research has shifted toward \emph{Iterative RAG}~\cite{feng2024retrieval,lee2024planrag}, which refines evidence collection across multiple turns, and explicit problem decomposition~\cite{ammann2025question} to handle multi-hop queries. Beyond flat retrieval, recent works attempt to incorporate structure. A first line constructs external hierarchies: \citet{sarthi2024raptor} and \citet{tao2025treerag} organize documents into tree-style representations via clustering or recursive summarization to support coarse-to-fine access. Another line models dependencies with graph abstractions, such as BookRAG~\cite{wang2025bookrag} and SentGraph~\cite{liang2026sentgraph}. While effective in some settings, these \emph{constructed} structures can diverge from the document’s \emph{native} layout and reading order, incurring non-trivial construction overhead and potentially disrupting the author’s intended narrative flow. In contrast, prior DocQA literature has long emphasized \emph{hierarchy- and order-aware} reading for long documents: \citet{choi2016hierarchical} proposes hierarchical decomposition to enable efficient reasoning over extended contexts, while \citet{mcdonald2022detect} formalizes a flexible detect--retrieve--comprehend pipeline that explicitly separates localization from comprehension. More recently, PDFTriage~\cite{saad2024pdftriage} demonstrates practical QA over long, structured PDFs, highlighting the importance of recovering reading order and consuming coherent regions rather than isolated snippets. Complementing these, recent surveys summarize the growing consensus that long-document retrieval must better respect document structure and sequential cues~\cite{li2025survey}. We follow this direction and argue that leveraging \emph{native} structural priors (headings and sequence) provides a more faithful and efficient interface for closed DocQA; \emph{DeepRead} operationalizes these priors via a coordinate-based navigation scheme that supports a human-like ``locate-then-read'' pattern.

\textbf{Document Parsing.} Reliable parsing serves as the bridge between raw visual documents and LLM reasoning. Recent advances in OCR have been transformative, with both pipeline systems (e.g., PaddleOCR-VL~\cite{cui2025paddleocr}) and end-to-end vision-language models (e.g., DeepSeek-OCR~\cite{wei2025deepseek}, HunyuanOCR~\cite{team2025hunyuanocr}) achieving high-fidelity results. Crucially, modern parsers can now output structured formats like Markdown or \LaTeX, effectively recovering not just characters but also the logical organization—headers, lists, and reading order. This breakdown of the barrier between visual layout and textual logic allows systems to treat parsed documents as structured artifacts rather than unordered bags of words. DeepRead leverages these high-quality parsers to construct a coordinate-based navigation system, enabling agents to perceive and traverse the document's native structure.

\textbf{Agentic Search.} Unlike static RAG pipelines, agentic RAG frames question answering as an autonomous decision-making process. Frameworks such as ReAct and the reasoning-centric Search-o1~\cite{li2025search} empower LLMs to dynamically decide \emph{when} to retrieve, \emph{what} to query, and \emph{how} to synthesize evidence, yielding strong performance on complex tasks. Despite this autonomy, a core limitation persists: most agentic approaches still treat documents as \emph{flat, unstructured collections of chunks}. This ``structural blindness'' obscures the logical position of evidence and fragments long-range context. We also note PageIndex~\cite{zhang2025pageindex}, a closed-source commercial system that emphasizes structure-only navigation based on pages and sections. Our observations and discussion of PageIndex are necessarily limited to publicly available materials and our brief front-end trial experience; due to limited information accessibility, we do not attempt a deeper technical analysis. Conceptually, relying solely on structural cues may sacrifice the rapid semantic positioning afforded by retrieval-based localization, a trade-off that we further validate in Table~\ref{tab:deepread_vs_read}. In contrast, \textbf{DeepRead} introduces a \emph{structure-grounded} agent that balances these extremes: it operationalizes document hierarchy as a lightweight coordinate system, using retrieval-assisted localization to quickly pinpoint relevant regions while employing contiguous, section-wise reading to support sustained, order-preserving reasoning.
\section{Methodology}
\label{sec:method}

This section defines (i) how we represent hierarchical document structure in a compact, addressable form and (ii) the formal interfaces of DeepRead's tools. Our goal is to enable an agent to navigate long documents with a human-like \emph{locate-then-read} pattern: first localize evidence efficiently, then read coherently in-order within the appropriate section.

\subsection{Preliminaries: Agentic Search}
\label{sec:preliminaries}

We adopt vanilla ReAct~\cite{yao2022react} as the agentic framework that interleaves reasoning and acting.
Let the user question be $q$. The interaction proceeds for at most $T$ rounds.
At round $t$, the agent state $s_t$ is the message history, including the system prompt, the user query, and the trajectory of tool interactions.
The model policy $\pi_\theta$ samples an action $a_t$:
\begin{equation}
a_t \sim \pi_\theta(\cdot \mid s_t),
\quad
a_t \in \{\textsf{FINAL}\} \cup \mathcal{A},
\end{equation}
where an action is either a final answer or a tool invocation.
A tool invocation is represented as $a_t=(\tau_t, \mathbf{x}_t)$ with tool name $\tau_t$ and arguments $\mathbf{x}_t$.
Executing $a_t$ yields an observation $o_t$, and the state is updated by appending the interaction:
\begin{equation}
s_{t+1} \leftarrow s_t \oplus (a_t, o_t),
\end{equation}
where $\oplus$ denotes concatenation to the history.
In DeepRead, the action set $\mathcal{A}$ consists of two tools, \textsf{Retrieve} and \textsf{ReadSection} (Sec.~\ref{sec:tools}),
which together support a human-like \emph{locate-then-read} workflow over structured documents.

\subsection{Document Structure Modeling}
\label{sec:modeling}

We assume raw documents are processed by an OCR engine and converted into a structured Markdown-like format.
To support precise navigation, we model document structure along two dimensions: hierarchy and sequence.
Concretely, we distinguish \emph{headings} (which define the hierarchy) and \emph{content paragraphs} (which define the reading sequence inside each heading).

\textbf{Entities and Metadata.}
We treat both headings and paragraphs as first-class \emph{entities}:
\begin{equation}
e = \big(t(e), \Gamma(e)\big),
\end{equation}
where $t(e)$ is the textual content and $\Gamma(e)$ is a structured metadata object.
This unified view is important in DeepRead: headings $h$ are used to build the global navigation map in the system prompt, while paragraphs $p$ are returned by tools as evidence together with their coordinates.

\textbf{Heading Entities.}
For a document $d$, let $N_h^{(d)}$ be the total number of headings.
We denote the $i$-th heading (in document order) as a heading entity
\begin{equation}
h^{(d)}_i = \big(t^{(d)}_i, \Gamma^{(d)}_i\big),
\quad i\in\{1,\dots,N_h^{(d)}\},
\end{equation}
where $t^{(d)}_i$ is the heading text and $\Gamma^{(d)}_i$ provides structural metadata. In particular, we define
\begin{equation}
\Gamma^{(d)}_i =
\left\{
\begin{aligned}
& \texttt{doc\_id}: d,\ \texttt{sec\_id}: i,\ \texttt{children}: C^{(d)}_i,\ \\
& \texttt{n\_para}: n^{(d)}_i,\ \texttt{n\_tok}: m^{(d)}_i
\end{aligned}
\right\}.
\label{eq:heading_meta}
\end{equation}
Here $C^{(d)}_i=\{k \mid \text{parent}(h^{(d)}_k)=h^{(d)}_i\}$ is the set of IDs of immediate children headings,
$n^{(d)}_i$ is the number of content paragraphs directly under heading $i$ (excluding its sub-headings),
and $m^{(d)}_i$ is the token count of these direct paragraphs.
This metadata allows the agent to infer both the nesting structure and the approximate reading cost of each section.

\textbf{Paragraph Entities.}
Within each heading $h^{(d)}_i$, we define its \emph{direct} content (excluding all sub-headings) as an ordered sequence of paragraphs:
\begin{equation}
P\left(h^{(d)}_i\right) =
\left[p^{(d)}_{i,1},\, p^{(d)}_{i,2},\, \dots,\, p^{(d)}_{i,n^{(d)}_i}\right],
\end{equation}
where $n^{(d)}_i$ matches the \texttt{n\_para} field in Eq.~\ref{eq:heading_meta}.
Each paragraph is also an entity:
\begin{equation}
p^{(d)}_{i,j} = \left(t^{(d)}_{i,j}, \Gamma^{(d)}_{i,j}\right),
\quad j\in\{1,\dots,n^{(d)}_i\},
\end{equation}
where $t^{(d)}_{i,j}$ is the paragraph text and the metadata encodes its coordinate:
\begin{equation}
\Gamma^{(d)}_{i,j} =
\left\{
\texttt{doc\_id}: d,\ 
\texttt{sec\_id}: i,\ 
\texttt{para\_idx}: j
\right\}.
\label{eq:para_meta}
\end{equation}
Thus, every paragraph is \emph{addressable} by $(d,i,j)$, i.e., \texttt{doc\_id}, \texttt{sec\_id}, and \texttt{para\_idx}.

Building upon the above modeling, DeepRead indexes atomic paragraphs as the fundamental retrieval units,
rather than merging text into arbitrary sliding windows.
During tool interaction, DeepRead always returns paragraph text together with $\Gamma^{(d)}_{i,j}$ (Eq.~\ref{eq:para_meta}).
This enables the agent to reason jointly about \emph{content} and \emph{location} (e.g., ``doc $d$, section $i$, paragraph $j$''),
which is essential for coordinate-based follow-up reading.

\subsection{Hierarchical Structure in System Prompt}
\label{sec:system_prompt}

To enable global planning without overwhelming the context window, DeepRead injects a lightweight structural representation
of the document collection into the system prompt.
Instead of providing full content, we serialize a compact Table of Contents (TOC) built from \emph{heading entities}.

For each document $d$, and for a collection of documents $\mathcal{D}$, we define
\begin{equation}
\begin{aligned}
\texttt{TOC}(d) &= \big[\, h^{(d)}_i \,\big]_{i=1}^{N_h^{(d)}}, \qquad
\texttt{TOC}(\mathcal{D}) &= \big[\, \texttt{TOC}(d) \,\big]_{d\in\mathcal{D}}.
\end{aligned}
\end{equation}

This design provides structural priors for planning:
by reading \texttt{children} in $\Gamma^{(d)}_i$, the agent can infer hierarchy and scope;
by reading \texttt{n\_para} and \texttt{n\_tok}, the agent can estimate reading cost and decide whether to read an entire section or a targeted span.
A concrete example of this schema is illustrated in Figure~\ref{fig:main-deepread}.

\begin{algorithm}[t]
\caption{DeepRead: Structure-Preserving Agentic Reading}
\label{alg:deepread}
\small
\begin{algorithmic}[1]
\Require Documents $\mathcal{D}$ with parsed headings $\{h^{(d)}_i\}$ and paragraphs $\{p^{(d)}_{i,j}\}$; Question $q$; Window $W=(w^{\uparrow}, w^{\downarrow})$.
\State \textbf{Initialize:} Construct system prompt with $\texttt{TOC}(\mathcal{D})$ (grouped by document; each $h^{(d)}_i$ includes $\Gamma^{(d)}_i$).
\State $s_1 \gets [\text{System},\ \text{User}: q]$
\For{$t = 1$ \textbf{to} $T$}
    \State $a_t \sim \pi_\theta(\cdot \mid s_t)$
    \If{$a_t = \textsf{FINAL}$} \Return Answer \EndIf

    \If{$a_t.\tau = \textsf{Retrieve}$}
        \State $u \gets a_t.\mathbf{x}.\texttt{query}$
        \State $H \gets \textsf{Rank}(u)$
        \Comment{$H=[(d_r,i_r,j_r,s_r)]_{r=1}^K$ sorted by $s_r$ descending}
        \State $\mathcal{U} \gets [\ ]$ \Comment{an ordered list of paragraph coordinates}
        \State $\mathcal{S} \gets \emptyset$ \Comment{a set for deduplication}
        \For{each hit $(d_r,i_r,j_r,s_r) \in H$}
            \State $j_r^{\uparrow} \gets \max(1,\ j_r-w^{\uparrow})$
            \State $j_r^{\downarrow} \gets \min(n^{(d_r)}_{i_r},\ j_r+w^{\downarrow})$
            \For{$j = j_r^{\uparrow}$ \textbf{to} $j_r^{\downarrow}$}
                \If{$(d_r,i_r,j)\notin \mathcal{S}$}
                    \State $\mathcal{U} \gets \mathcal{U} \oplus [(d_r,i_r,j)]$
                    \State $\mathcal{S} \gets \mathcal{S} \cup \{(d_r,i_r,j)\}$
                \EndIf
            \EndFor
        \EndFor
        \State $o_t \gets \textsf{Format}\Big(\big\{p^{(d)}_{i,j} : (d,i,j)\in \mathcal{U}\big\}\Big)$
        \Comment{\textsf{Format} serializes paragraphs in the list order and preserves $\Gamma^{(d)}_{i,j}$}
    \ElsIf{$a_t.\tau = \textsf{ReadSection}$}
        \State $d \gets a_t.\mathbf{x}.\texttt{doc\_id}$; $i \gets a_t.\mathbf{x}.\texttt{sec\_id}$
        \State $j_s \gets a_t.\mathbf{x}.\texttt{start}$; $j_e \gets a_t.\mathbf{x}.\texttt{end}$
        \State $j_s \gets \max(1,\ j_s)$; $j_e \gets \min(n^{(d)}_{i},\ j_e)$
        \Comment{Clip to valid range using $n^{(d)}_{i}$ from the TOC metadata}
        \State $o_t \gets \textsf{Format}\Big(\big\{p^{(d)}_{i,j} : j\in[j_s,j_e]\big\}\Big)$
        \Comment{\textsf{Format} returns contiguous paragraphs in increasing $j$ with metadata}
    \EndIf

    \State $s_{t+1} \leftarrow s_t \oplus (a_t, o_t)$
\EndFor
\end{algorithmic}
\end{algorithm}

\subsection{Tools: Coordinate-Based Interaction}
\label{sec:tools}

The agent interacts with the document collection via two complementary tools defined on the paragraph coordinate system $(d,i,j)$.
This interaction mimics human behavior: \emph{fast localization} via retrieval, followed by \emph{order-preserving reading} within the appropriate section.

\textbf{Output Convention: \textsf{Format}.}
Both tools return \emph{paragraph entities} $p^{(d)}_{i,j}=(t^{(d)}_{i,j},\Gamma^{(d)}_{i,j})$.
In practice, \textsf{Format} serializes each paragraph by prefixing (or otherwise attaching) its metadata $\Gamma^{(d)}_{i,j}$ (Eq.~\ref{eq:para_meta}),
so the agent can cite and navigate using explicit coordinates.

\noindent\textbf{1.\ \textsf{Retrieve}.}
\textsf{Retrieve} is a locator tool that accepts a query string $u$ (via the \texttt{query} field).
It performs semantic retrieval over paragraph entities and returns top-$K$ hits with scores:
\begin{equation}
\textsf{Rank}(u) \rightarrow \big[(d_r,i_r,j_r,s_r)\big]_{r=1}^K,
\quad \text{with } s_1 \ge s_2 \ge \cdots \ge s_K.
\end{equation}
To simulate human skimming (inspecting nearby context), we introduce a scanning window
$W=(w^{\uparrow}, w^{\downarrow})$ representing upward and downward look-ahead sizes.
For a hit at coordinate $(d_r,i_r,j_r)$, the scan boundaries are
\begin{equation}
j_r^{\uparrow} = \max(1,\ j_r - w^{\uparrow}),
\quad
j_r^{\downarrow} = \min(n^{(d_r)}_{i_r},\ j_r + w^{\downarrow}),
\end{equation}
yielding the local slice
\begin{equation}
\textsf{Scan}(d_r,i_r,j_r;W) = \big[(d_r,i_r,j)\big]_{j=j_r^{\uparrow}}^{j_r^{\downarrow}},
\end{equation}
where paragraphs are ordered by $j$ within each slice.
DeepRead expands \emph{each} hit independently following the ranked order of $H$ and deduplicates overlaps
\emph{while preserving first-occurrence order}:
\begin{equation}
\mathcal{U}(u) = \textsf{Unique}\Big(\ \bigoplus_{r=1}^{K}\ \textsf{Scan}(d_r,i_r,j_r;W)\ \Big),
\label{eq:ranked_unique}
\end{equation}
where $\oplus$ denotes list concatenation, and $\textsf{Unique}(\cdot)$ removes repeated coordinates by keeping the first occurrence.

Finally, \textsf{Retrieve} returns the corresponding paragraph entities in the list order:
\begin{equation}
\textsf{Retrieve}(u) \rightarrow
\textsf{Format}\Big(\big[p^{(d)}_{i,j}\big]_{(d,i,j)\in \mathcal{U}(u)}\Big).
\end{equation}

\noindent\textbf{2.\ \textsf{ReadSection}.}
\textsf{ReadSection} performs deep, order-preserving reading over a targeted region.
It accepts a document ID $d$, a section ID $i$ (unique within $d$), and a paragraph range $[j_{\text{start}}, j_{\text{end}}]$:
\begin{equation}
\textsf{ReadSection}(d,i,j_{\text{start}},j_{\text{end}})
\rightarrow
\textsf{Format}\Big(\{p^{(d)}_{i,j} : j\in[j_{\text{start}},j_{\text{end}}]\}\Big).
\end{equation}
The system clips the requested range to valid boundaries using $n^{(d)}_i$ in heading metadata $\Gamma^{(d)}_i$ (Eq.~\ref{eq:heading_meta}).
This returns contiguous, in-order paragraphs from the specified section, reducing context fragmentation introduced by retrieval.

\textbf{Synergy.}
The two tools form a closed loop.
\textsf{Retrieve} provides (i) a lightweight preview and (ii) coordinate anchors $(d,i,j)$ for relevant evidence.
When the agent determines that additional context is needed (e.g., preceding/following paragraphs or a larger span within the same section),
it invokes \textsf{ReadSection}.
Together, these tools enable faithful long-document reasoning via a human-like \emph{locate-then-read} paradigm, as summarized in Algorithm~\ref{alg:deepread}.
\begin{table*}[!t]
    \centering
    \caption{Comparison with Different Methods (Accuracy \%). \textbf{Bold} indicates the optimal choice, \underline{underlined} indicates the next best choice. Green text denotes the absolute improvement over the corresponding Search-o1 baseline.}
    \label{tab:method_comparison}

    \resizebox{\linewidth}{!}{
        \begin{tabular}{lccccccc}
            \toprule
            \multirow{2}{*}{\textbf{Method}} & \multicolumn{3}{c}{\textbf{Single-Document}} & \multicolumn{3}{c}{\textbf{Multi-Document}} & \multirow{2}{*}{\textbf{Overall Avg}} \\
            \cmidrule(lr){2-4} \cmidrule(lr){5-7}
             & \textbf{FinanceBench~\cite{islam2023financebench}} & \textbf{ContextBench (Ours)} & \textbf{Avg} & \textbf{QASPER~\cite{dasigi2021dataset}} & \textbf{SyllabusQA~\cite{fernandez2024syllabusqa}} & \textbf{Avg} & \\
            \midrule
            Dense RAG~\cite{lewis2020retrieval,qwen3embedding}            & 38.1 & 60.0 & 49.0 & 15.4 & 26.3 & 20.8 & 34.9 \\
            Dense RAG w/ Reranker~\cite{qwen3embedding}       & 47.3 & 66.2 & 56.8 & 15.4 & 24.8 & 20.1 & 38.4 \\
            ITRG (refresh)~\cite{feng2024retrieval}           & 48.1 & 67.9 & 58.0 & 9.8 & 31.0 & 20.4 & 39.2 \\
            ITRG (refine)~\cite{feng2024retrieval}            & 52.0 & 70.1 & 61.0 & 15.4 & 31.3 & 23.3 & 42.2 \\
            RAPTOR~\cite{sarthi2024raptor} & 38.7 & 52.6 & 45.7 & 20.2 & 40.3 & 30.3 & 38.0 \\
            \hline
            Search-o1~\cite{li2025search}                      & 80.0 & 74.5 & 77.3 & 65.0 & 57.1 & 61.1 & 69.2 \\
            DeepRead (Ours)                                    & 82.7 \rise{2.7} & \textbf{91.5} \rise{17.0} & \textbf{87.1} \rise{9.8} & \underline{72.7} \rise{7.7} & \underline{70.9} \rise{13.8} & \underline{71.8} \rise{10.7} & \underline{79.5} \rise{10.3} \\
            \hline
            Search-o1 w/ expand~\cite{li2025search}             & \underline{83.3} & 84.0 & 83.7 & 65.0 & 68.4 & 66.7 & 75.2 \\
            DeepRead w/ expand (Ours)                          & \textbf{84.0} \rise{0.7} & \underline{88.3} \rise{4.3} & \underline{86.2} \rise{2.5} & \textbf{76.2} \rise{11.2} & \textbf{72.4} \rise{4.0} & \textbf{74.3} \rise{7.6} & \textbf{80.3} \rise{5.1} \\
            \bottomrule
        \end{tabular}
    }
\end{table*}

\label{sec:deepread_algo}

\section{Experiment}

\subsection{Benchmark Details}
We evaluated DeepRead on four benchmarks designed to test specific RAG capabilities: 

\textbf{(1) FinanceBench~\cite{islam2023financebench}:} This is a long-document QA benchmark for the financial sector, with documents sourced from SEC financial filings. We use the open-source version (150 pairs) to evaluate long-document reasoning within the financial domain.

\textbf{(2) ContextBench (Ours):} This benchmark specifically collects QA tasks on long documents that require extensive context or long-range dependencies. This benchmark comprises 12 AI experts using PDFs from their daily work and personal lives, including novels, academic papers, scripts, textbooks, etc. These experts provided real-world questions, with each question taking approximately 0.5 man-hours to annotate, resulting in 94 QA pairs  (details are provided in Appendix~\ref{ap:bench_detail}).

\textbf{(3) QASPER~\cite{dasigi2021dataset} (Multi-Doc):} To test academic and cross document reasoning, we synthesized a multi-document version of QASPER. We used an LLM to generate questions spanning 2--5 papers and manually filtered out illogical or erroneous samples, resulting in 143 high-quality pairs (details are provided in Appendix~\ref{ap:bench_detail} and Figure~\ref{fig:qa_gen_prompt}).

\textbf{(4) SyllabusQA~\cite{fernandez2024syllabusqa} (Multi-Doc):} To test documents with simple hierarchical structures, we obtained all course syllabus PDFs from SyllabusQA for constructing single-document QA, we applied the same synthesis and manual verification process as QASPER, yielding 196 high-quality pairs. (details are provided in Appendix~\ref{ap:bench_detail} and Figure~\ref{fig:qa_gen_prompt})

Table~\ref{tab:dataset_statistics} presents the relevant statistical data for the above benchmarks. Regarding the construction of synthetic multi-document benchmarks, we utilized GLM-4.7~\cite{5team2025glm45agenticreasoningcoding} taking the full document context as input, with a temperature of 0.7 to encourage diversity in question generation. Document structure parsing was conducted using PaddleOCR-VL, served via VLLM~\cite{kwon2023efficient} on an NVIDIA RTX 4090 GPU. We focus on end-to-end performance using an LLM-as-a-Judge framework. We employ DeepSeek V3.2~\cite{deepseekai2025deepseekv32} as the evaluator, with the temperature set to 0.0 reduce variance. Due to differing chunking strategies across methods, metrics such as intermediate chunk recall cannot be fairly compared. To evaluate the correctness of final answers, we employ DeepSeek V3.2 for judgment. Prompts are shown in Figure ~\ref{fig:judge_prompt}.

\begin{figure}[t]
  \centering
  \includegraphics[width=\columnwidth]{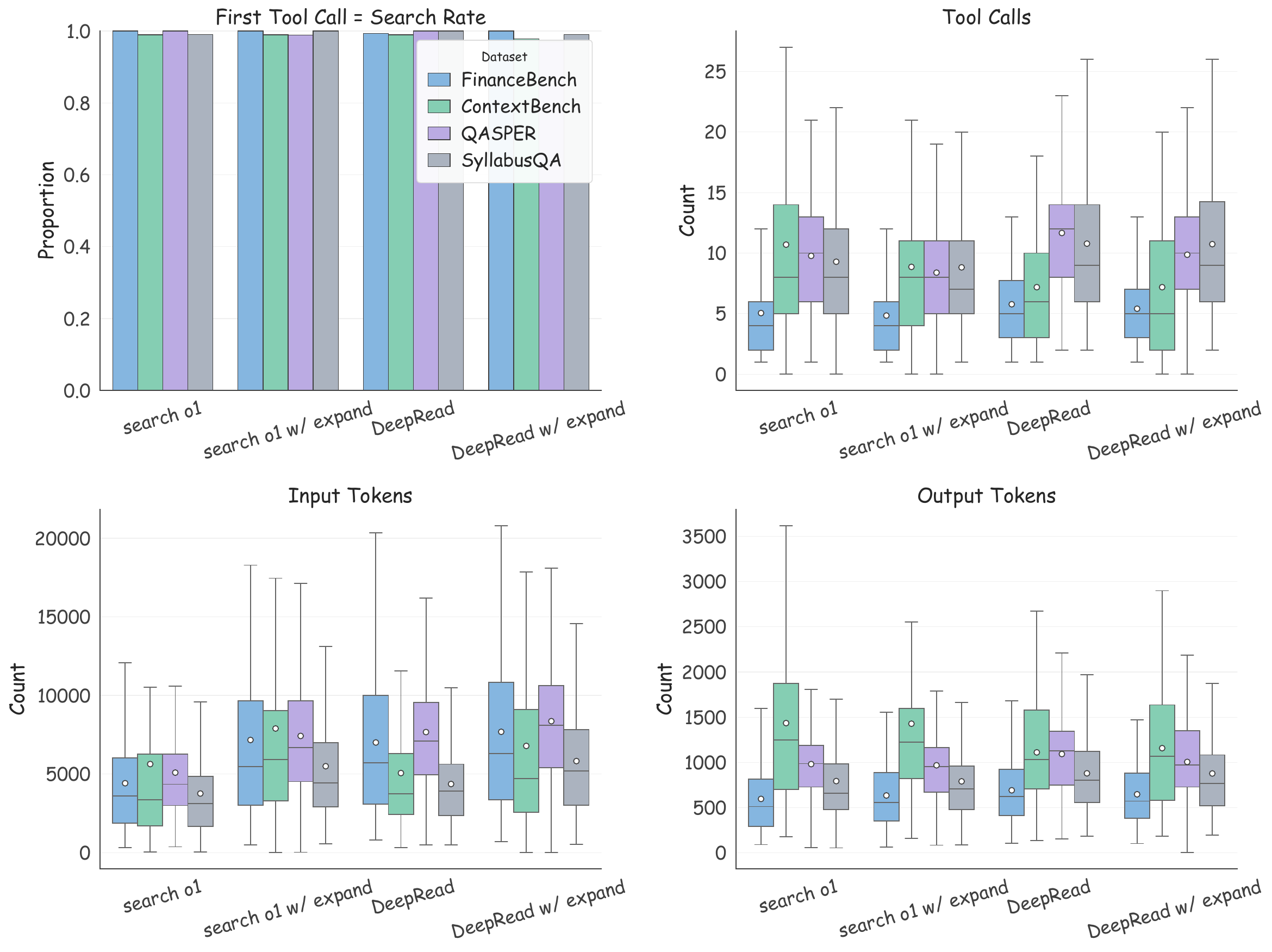}
  \caption{\textbf{Fine-grained behavioral comparison between DeepRead and Search-o1 baselines.} We illustrate the distribution of the probability that the first action is a search, the total number of tool calls per query, input token consumption, and output token generation across four benchmarks.}
  \label{fig:Fine}
\end{figure}
\subsection{Baseline and DeepRead Settings}
To evaluate the effectiveness of DeepRead, we consider four baseline families: single-pass retrieval, RAPTOR~\cite{sarthi2024raptor}, Iterative Retrieval Generation Synergy (ITRG)~\cite{feng2024retrieval}, and Search-o1~\cite{li2025search}. We did not compare methods such as Search-R1~\cite{jin2025search} that are dedicated to training agents, but these methods all employ a retrieval tool for ReAct inference.

We use Qwen3-embedding-8b~\cite{qwen3embedding} as the dense retriever and Qwen3-reranker-8b~\cite{qwen3embedding} as the reranker. Since two-stage retrieval is the de facto industrial practice, we apply a reranker by default except RAPTOR as it cannot apply reranking due to its method design. Moreover, we also explore the result of single-pass retrieval without a reranker.

Concretely, for single-pass and ITRG, we follow the OpenAI File Search configuration with chunk size 800 and overlap 400. For \textbf{single-pass retrieval} return the top 10 chunks in one round (as in Search-o1 paper). When reranking is enabled, the first-stage retriever produces 30 candidates, which are then scored by a reranker and truncated to the target token budget. For \textbf{RAPTOR}, we utilize the recommended ``Collapsed Tree'' setting~\cite{sarthi2024raptor} with a maximum token limit of 800 per node, 5 layers, and a clustering top-$k$ of 5, retrieving the top-10 nodes from the collapsed index. For \textbf{ITRG}, we adopt the more effective 4-round setting reported in the original paper, returning the top 6 chunks per round. For \textbf{Search-o1 and our method}, they uses structure-based chunking with overlap 0, and each retrieval tool call returns 2 chunks that appears relatively small, but due to multiple rounds, the total number of tokens after multiple rounds is comparable to other methods. When applying context expansion, the expansion window is configured as (1, 1). We did not expand further because extending a paragraph by one segment effectively triples the number of individual retrievals, which would weaken the semantic retrieval capability. For Search-o1, to ensure fair comparison, we do not include the Reason-in-Documents, and we additionally inject the document structural schema into the system prompt (matching DeepRead’s access to structure). We excluded the 'Reason-in-Documents' module because it functions as a high-cost optimization—requiring repetitive summarization of interaction history—that yields only marginal performance gains. As this technique is an orthogonal enhancement applicable to general ReAct paradigms rather than a core architectural feature, its omission aligns with subsequent agentic search methodologies (e.g., Search R1~\cite{jin2025search}) that prioritize efficiency, thereby ensuring a fair and representative comparison. 

The policy model of all baseline and DeepRead used in this paper is DeepSeek v3.2, with a decoding temperature of 0. For search-o1 and DeepRead, we set the maximum round to 50.

\subsection{Main Result}
\label{sec:main_result}

Table~\ref{tab:method_comparison} reports end-to-end accuracy on four benchmarks.
Across all settings, \textbf{DeepRead consistently outperforms strong baselines}, with the largest gains on ContextBench, which requires long-range, scope-aware evidence integration.
DeepRead achieves an \textbf{overall average of 79.5\%} (80.3\% with \texttt{expand}), outperforming Search-o1 by \textbf{+10.3} points (and by \textbf{+5.1} points under \texttt{expand}).
These results suggest that explicitly exposing document hierarchy and sequential structure to the agent yields substantial benefits over treating long documents as flat, orderless chunks.

\textbf{Impact of structure-grounded reading (\texttt{ReadSection}).}
Comparing Search-o1 and DeepRead isolates the contribution of coordinate-based, order-preserving reading.
DeepRead improves upon Search-o1 on \textbf{all} benchmarks, with especially large gains on \textbf{ContextBench} (\textbf{+17.0} points; 74.5\% $\rightarrow$ 91.5\%).
This supports our central claim: flat retrieval fragments discourse and forces the agent to stitch evidence from disjoint chunks, whereas \texttt{ReadSection} reconstructs \emph{contiguous} evidence anchored to explicit structural coordinates.
Notably, DeepRead also yields strong improvements in multi-document settings, achieving \textbf{+7.7} on QASPER (65.0\% $\rightarrow$ 72.7\%) and \textbf{+13.8} on SyllabusQA (57.1\% $\rightarrow$ 70.9\%),
indicating that hierarchy- and sequence-aware navigation remains effective even when evidence spans multiple files and sections.

\textbf{Dynamic reading vs.\ passive expansion (\texttt{ReadSection} vs.\ \texttt{expand}).}
Both \texttt{ReadSection} and \texttt{expand} enlarge context, but they do so in fundamentally different ways.
\texttt{expand} is a \emph{passive, structure-only} heuristic: it blindly appends nearby paragraphs around retrieved hits, regardless of whether they are semantically useful.
In contrast, \texttt{ReadSection} is \emph{dynamic and semantics-grounded}: the agent first localizes evidence and then selectively reads a coherent span in the most relevant section.
The results in Table~\ref{tab:method_comparison} reflect this distinction.
Even without \texttt{expand}, DeepRead (79.5\%) markedly exceeds Search-o1 (69.2\%), showing that semantic, coordinate-based reading is more effective than relying on window-based padding to mitigate fragmentation.

\textbf{Effect of local expansion (\texttt{expand}).}
Expansion generally benefits retrieval-heavy baselines, but its effect is not uniformly positive for structure-grounded reading.
Search-o1 improves substantially with \texttt{expand} (69.2\% $\rightarrow$ 75.2\%), as adjacent context partially compensates for the \emph{context fragmentation} inherent in flat retrieval.
In contrast, DeepRead exhibits only a modest overall gain (79.5\% $\rightarrow$ 80.3\%).
Moreover, on ContextBench, \texttt{expand} \emph{reduces} DeepRead accuracy (91.5\% $\rightarrow$ 88.3\%).
We manually inspected failure cases introduced by \texttt{expand} for our method and found that the dominant issue is that \texttt{expand} brings in paragraphs irrelevant to the target answer, which in turn misleads the agent into issuing incorrect queries or invoking inappropriate tools.
Overall, these results suggest that both \texttt{ReadSection} and \texttt{expand} can supplement context in a structure-aware manner, but \texttt{ReadSection} does so more precisely and effectively.

\textbf{Robustness to judge choice.}
To reduce dependence on a single evaluator, we replicate all experiments using two additional independent LLM judges, GLM-4.7~\cite{5team2025glm45agenticreasoningcoding} and Qwen3-235B-A22B-thinking-2507~\cite{qwen3technicalreport} (Appendix~\ref{app:judge}).
Across judges, the \textbf{relative ranking is stable}: DeepRead remains consistently stronger than Search-o1 variants, indicating that the gains are not an artifact of any single judge’s calibration.
Furthermore, the inter-judge agreement reported in Table~\ref{tab:judge_agreement} is high overall, supporting the reliability of the observed improvements.
\begin{table}[t]
\centering
\caption{The proportion of samples whose first tool call is \texttt{Retrieve} and use \texttt{Read} subsequently ($S_{s \to r}$), and the ratio of calls of \texttt{Retrieval} to \texttt{Read} ($C_s/C_r$).}
\resizebox{\columnwidth}{!}{%
\begin{tabular}{lcccc}
\toprule
 & \multicolumn{2}{c}{\textbf{DeepRead}} & \multicolumn{2}{c}{\textbf{DeepRead w/ expand}} \\
\cmidrule(lr){2-3} \cmidrule(lr){4-5}
\textbf{Benchmark} & $S_{s \to r}$ (\%) & \textbf{$C_s/C_r$} & $S_{s \to r}$ (\%) & \textbf{$C_s/C_r$} \\
\midrule
ContextBench & 95.70\% & 0.87 & 93.55\% & 0.88 \\
FinanceBench & 87.33\% & 1.82 & 82.67\% & 2.18 \\
QASPER       & 98.25\% & 1.59 & 92.40\% & 2.00 \\
SyllabusQA   & 97.96\% & 1.61 & 96.43\% & 1.86 \\
\bottomrule
\end{tabular}%
}
\label{tab:search_read_stats}
\end{table}
\subsection{Fine-Grained Behavior Analysis}
\label{sec:behavior_analysis}

The preceding quantitative results demonstrate the performance superiority of DeepRead. Here, we conduct a fine-grained behavioral analysis characterizing how DeepRead diverges from standard Search-o1-style agentic workflows in terms of planning, tool consumption, and information processing efficiency.

Figure~\ref{fig:Fine} visualizes the distribution of agent behaviors across four experimental settings, and Table~\ref{tab:search_read_stats} quantifies a recurring pattern that \emph{emerges autonomously} in DeepRead rather than being manually hard-coded as a fixed workflow. Specifically, $S_{s \to r}$ measures the fraction of questions where the agent starts with \texttt{Retrieve} and later invokes \texttt{Read}, while $C_s/C_r$ summarizes the balance between localization (\texttt{Retrieve}) and contiguous reading (\texttt{Read}). As shown in Table~\ref{tab:search_read_stats}, $S_{s \to r}$ is consistently high across all benchmarks (87.33\%--98.25\% without \texttt{expand}, 82.67\%--96.43\% with \texttt{expand}), indicating that the policy model \emph{learns to} first obtain coordinate anchors via retrieval and then switch to section-wise reading for evidence consolidation, instead of answering from fragmented snippets. Moreover, $C_s/C_r$ adapts to task characteristics: ContextBench is more read-heavy ($C_s/C_r \approx 0.87$), consistent with section-scoped evidence, whereas FinanceBench is more retrieval-heavy ($C_s/C_r = 1.82$), reflecting the need to pinpoint specific tables or numeric fields before reading. Enabling \texttt{expand} generally increases $C_s/C_r$ (e.g., 1.82$\rightarrow$2.18 on FinanceBench and 1.59$\rightarrow$2.00 on QASPER), suggesting that expansion partially substitutes for deep reading by enriching retrieval outputs, while the persistently high $S_{s \to r}$ confirms that DeepRead still predominantly exhibits this emergent locate-then-read behavior.

Further examination of resource consumption metrics (Figure~\ref{fig:Fine} and Table~\ref{tab:cost_correct_vs_wrong}) reveals that DeepRead incurs a higher computational overhead compared to the baseline. However, we argue that this reflects a favorable \textit{cost-performance trade-off}. The baseline's lower consumption stems from ``context starvation,'' where skipping necessary reading leads to significantly lower accuracy. DeepRead's ``locate-then-read'' paradigm invests in consuming contiguous sections to ensure \textit{informational sufficiency} for reasoning. As shown in Table~\ref{tab:method_comparison}, this moderate increase in token consumption yields a substantial \textbf{+10.3\%} improvement in overall accuracy. This indicates that the additional cost is not inefficiency, but a necessary investment for faithful long-document reasoning, avoiding the prohibitive costs associated with complex knowledge graph construction or multi-stage iterative summarization.

We also investigate the behavioral divergence between successful and failed queries. Our analysis reveals that incorrect samples frequently exhibit pathological search patterns characterized by prolonged tool usage, which results in a prohibitive escalation of resource consumption, as evidenced in Table~\ref{tab:cost_correct_vs_wrong}.
\begin{table}[t]
\centering
\caption{\textbf{Cost comparison between correct and incorrect samples.} We report the average number of tool calls and total token consumption across all benchmarks.}
\label{tab:cost_correct_vs_wrong}
\resizebox{\columnwidth}{!}{%
\begin{tabular}{lcccc}
\toprule
\multirow{2}{*}{\textbf{Method}} & \multicolumn{2}{c}{\textbf{Tool Calls}} & \multicolumn{2}{c}{\textbf{Total Tokens}} \\
\cmidrule(lr){2-3} \cmidrule(lr){4-5}
 & \textbf{Correct} & \textbf{Wrong} & \textbf{Correct} & \textbf{Wrong} \\
\midrule
Search-o1 & 7.7 & 10.4 & 4,978 & 6,522 \\
Search-o1 w/ expand & 7.1 & 9.5 & 7,349 & 8,700 \\
DeepRead & 8.7 & 11.0 & 6,770 & 7,648 \\
DeepRead w/ expand & 8.1 & 10.4 & 7,597 & 9,609 \\
\bottomrule
\end{tabular}%
}
\end{table}
\subsection{Ablation Study}
\begin{table}[t]
\centering
\caption{Performance and Cost Comparison between DeepRead and Readonly Baseline.}
\label{tab:deepread_vs_read}
\resizebox{\linewidth}{!}{%
\begin{tabular}{lcccccc}
\toprule
\multirow{2}{*}{\textbf{Benchmark}} & \multicolumn{3}{c}{\textbf{DeepRead}} & \multicolumn{3}{c}{\textbf{Readonly}} \\
\cmidrule(lr){2-4} \cmidrule(lr){5-7}
 & \textbf{Acc (\%)} & \textbf{Tools} & \textbf{Tokens} & \textbf{Acc (\%)} & \textbf{Tools} & \textbf{Tokens} \\
\midrule
FinanceBench & 82.7 & 5.8 & 7,698 & 80.7 & 4.7 & 7,980 \\
ContextBench & 91.5& 7.2 & 6,172 & 91.5& 6.9 & 6,680 \\
QASPER       & 72.7 & 11.7 & 8,768 & 18.9 & 9.9 & 9,887 \\
SyllabusQA   & 70.9 & 10.8 & 5,244 & 15.3 & 7.4 & 4,189 \\
\bottomrule
\end{tabular}%
}
\end{table}
Given that Table \ref{tab:method_comparison} has already demonstrated the results of whether to expand and only retrieve (search-o1), here we are primarily concerned with the synergistic effect of retrieval and read operations. Here, we conducted an ablation experiment to specifically evaluate the role and effectiveness of the Read tool within DeepRead. We found that in single-document scenarios, allowing the LLM to perform Read without retrieval is competitive in terms of efficiency and cost. However, in multi-document scenarios, it exhibits significant disadvantages in both efficiency and cost, as detailed in Table~\ref{tab:deepread_vs_read}. This validates the synergistic effect between retrieval and reading tools, particularly in multi-document contexts.

\begin{figure}[h]
    \centering
    \includegraphics[width=\columnwidth]{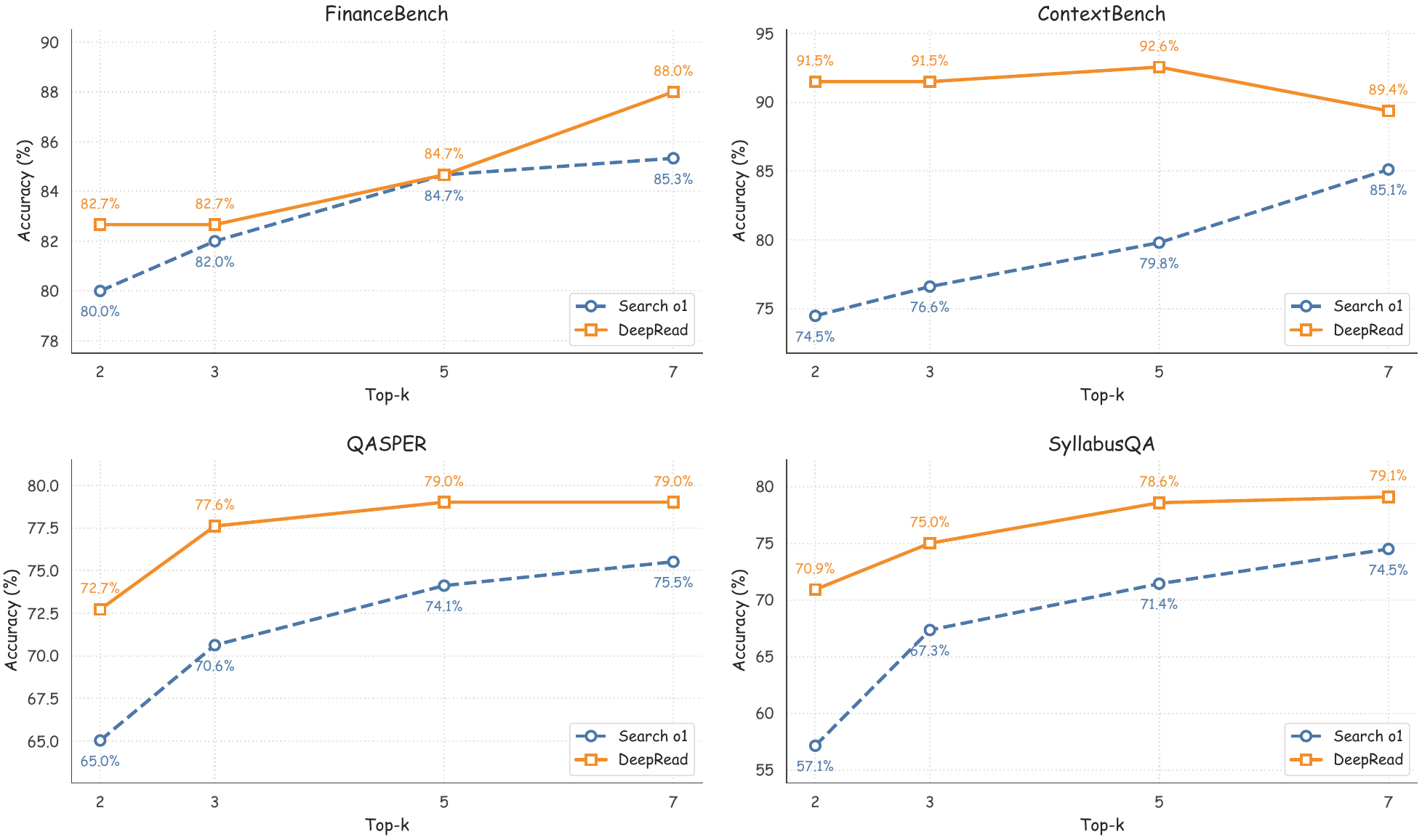} 
    \caption{\textbf{Impact of Retrieved Chunk Count ($k$) on Performance.} We compare DeepRead against Search-o1 across four benchmarks with $k \in \{2, 3, 5, 7\}$.}
    \label{fig:top_k_ablation}
\end{figure}

We further investigate the robustness of DeepRead with respect to the number of retrieved chunks ($k$). As illustrated in Figure~\ref{fig:top_k_ablation}, although higher retrieval recall generally correlates with improved accuracy, DeepRead consistently outperforms the Search-o1 baseline across all tested benchmarks and $k$ values. In many settings, Search-o1 is allowed to retrieve more chunks per round, yet it still underperforms DeepRead, which attains better accuracy with a smaller per-round retrieval set—suggesting that simply increasing the number of retrieved chunks in Search-o1 is insufficient to close the gap.

\section{Case Study}
We provide cases in Appendix~\ref{app:Case} which also demonstrate behaviors similar to human reading and searching.
\section{Conclusion}
This work introduces DeepRead, a structure-aware document reasoning agent for agentic RAG that mitigates the structural blindness of mainstream retrieval pipelines. DeepRead operationalizes document-native hierarchical and sequential priors as a  coordinate system, exposing two tools—Retrieve for scanning-based localization and ReadSection for contiguous, order-preserving reading within targeted scopes. This interface encourages an emergent locate-then-read strategy: the agent first anchors on relevant regions and then consolidates evidence by reading coherent spans, reducing context fragmentation and avoiding redundant, keyword-driven retrieval. Our experiments demonstrate consistent improvements over strong Search-o1-style baselines, and our fine-grained behavioral analyses show that DeepRead adopts human-aligned navigation patterns by balancing lightweight localization with selective deep reading. Overall, our results suggest that exposing and leveraging native document structure is a practical and effective step toward faithful, efficient reasoning over long, organized documents in agentic search.
\bibliographystyle{ACM-Reference-Format}
\bibliography{ref}

@article{lewis2020retrieval,
  title={Retrieval-augmented generation for knowledge-intensive nlp tasks},
  author={Lewis, Patrick and Perez, Ethan and Piktus, Aleksandra and Petroni, Fabio and Karpukhin, Vladimir and Goyal, Naman and K{\"u}ttler, Heinrich and Lewis, Mike and Yih, Wen-tau and Rockt{\"a}schel, Tim and others},
  journal={Advances in neural information processing systems},
  volume={33},
  pages={9459--9474},
  year={2020}
}

@inproceedings{feng2024retrieval,
  title={Retrieval-generation synergy augmented large language models},
  author={Feng, Zhangyin and Feng, Xiaocheng and Zhao, Dezhi and Yang, Maojin and Qin, Bing},
  booktitle={ICASSP 2024-2024 IEEE International Conference on Acoustics, Speech and Signal Processing (ICASSP)},
  pages={11661--11665},
  year={2024},
  organization={IEEE}
}

@article{li2025survey,
  title={A Survey of Long-Document Retrieval in the PLM and LLM Era},
  author={Li, Minghan and Luo, Miyang and Lv, Tianrui and Zhang, Yishuai and Zhao, Siqi and Nie, Ercong and Zhou, Guodong},
  journal={arXiv preprint arXiv:2509.07759},
  year={2025}
}

@article{choi2016hierarchical,
  title={Hierarchical question answering for long documents},
  author={Choi, Eunsol and Hewlett, Daniel and Lacoste, Alexandre and Polosukhin, Illia and Uszkoreit, Jakob and Berant, Jonathan},
  journal={arXiv preprint arXiv:1611.01839},
  year={2016}
}

@article{mcdonald2022detect,
  title={Detect, retrieve, comprehend: a flexible framework for zero-shot document-level question answering},
  author={McDonald, Tavish and Tsan, Brian and Saini, Amar and Ordonez, Juanita and Gutierrez, Luis and Nguyen, Phan and Mason, Blake and Ng, Brenda},
  journal={arXiv preprint arXiv:2210.01959},
  year={2022}
}

@inproceedings{saad2024pdftriage,
  title={Pdftriage: Question answering over long, structured documents},
  author={Saad-Falcon, Jon and Barrow, Joe and Siu, Alexa and Nenkova, Ani and Yoon, Seunghyun and Rossi, Ryan A and Dernoncourt, Franck},
  booktitle={Proceedings of the 2024 Conference on Empirical Methods in Natural Language Processing: Industry Track},
  pages={153--169},
  year={2024}
}

@article{liang2026sentgraph,
  title={SentGraph: Hierarchical Sentence Graph for Multi-hop Retrieval-Augmented Question Answering},
  author={Liang, Junli and Zhou, Pengfei and Zhou, Wangqiu and Qing, Wenjie and Zhao, Qi and Wang, Ziwen and Song, Qi and Li, Xiangyang},
  journal={arXiv preprint arXiv:2601.03014},
  year={2026}
}

@article{lee2024planrag,
  title={Planrag: A plan-then-retrieval augmented generation for generative large language models as decision makers},
  author={Lee, Myeonghwa and An, Seonho and Kim, Min-Soo},
  journal={arXiv preprint arXiv:2406.12430},
  year={2024}
}

@article{ammann2025question,
  title={Question Decomposition for Retrieval-Augmented Generation},
  author={Ammann, Paul JL and Golde, Jonas and Akbik, Alan},
  journal={arXiv preprint arXiv:2507.00355},
  year={2025}
}

@article{wang2025bookrag,
  title={BookRAG: A Hierarchical Structure-aware Index-based Approach for Retrieval-Augmented Generation on Complex Documents},
  author={Wang, Shu and Zhou, Yingli and Fang, Yixiang},
  journal={arXiv preprint arXiv:2512.03413},
  year={2025}
}

@article{zhao2024longrag,
  title={Longrag: A dual-perspective retrieval-augmented generation paradigm for long-context question answering},
  author={Zhao, Qingfei and Wang, Ruobing and Cen, Yukuo and Zha, Daren and Tan, Shicheng and Dong, Yuxiao and Tang, Jie},
  journal={arXiv preprint arXiv:2410.18050},
  year={2024}
}

@article{gong2025mmrag,
  title={Mmrag-docqa: A multi-modal retrieval-augmented generation method for document question-answering with hierarchical index and multi-granularity retrieval},
  author={Gong, Ziyu and Huang, Yihua and Mai, Chengcheng},
  journal={arXiv e-prints},
  pages={arXiv--2508},
  year={2025}
}

@article{islam2023financebench,
  title={Financebench: A new benchmark for financial question answering},
  author={Islam, Pranab and Kannappan, Anand and Kiela, Douwe and Qian, Rebecca and Scherrer, Nino and Vidgen, Bertie},
  journal={arXiv preprint arXiv:2311.11944},
  year={2023}
}

@article{dasigi2021dataset,
  title={A dataset of information-seeking questions and answers anchored in research papers},
  author={Dasigi, Pradeep and Lo, Kyle and Beltagy, Iz and Cohan, Arman and Smith, Noah A and Gardner, Matt},
  journal={arXiv preprint arXiv:2105.03011},
  year={2021}
}

@article{fernandez2024syllabusqa,
  title={SyllabusQA: A course logistics question answering dataset},
  author={Fernandez, Nigel and Scarlatos, Alexander and Lan, Andrew},
  journal={arXiv preprint arXiv:2403.14666},
  year={2024}
}

@article{qwen3embedding,
  title={Qwen3 Embedding: Advancing Text Embedding and Reranking Through Foundation Models},
  author={Zhang, Yanzhao and Li, Mingxin and Long, Dingkun and Zhang, Xin and Lin, Huan and Yang, Baosong and Xie, Pengjun and Yang, An and Liu, Dayiheng and Lin, Junyang and Huang, Fei and Zhou, Jingren},
  journal={arXiv preprint arXiv:2506.05176},
  year={2025}
}

@misc{qwen3technicalreport,
      title={Qwen3 Technical Report}, 
      author={Qwen Team},
      year={2025},
      eprint={2505.09388},
      archivePrefix={arXiv},
      primaryClass={cs.CL},
      url={https://arxiv.org/abs/2505.09388}, 
}

@inproceedings{yao2022react,
  title={React: Synergizing reasoning and acting in language models},
  author={Yao, Shunyu and Zhao, Jeffrey and Yu, Dian and Du, Nan and Shafran, Izhak and Narasimhan, Karthik R and Cao, Yuan},
  booktitle={The eleventh international conference on learning representations},
  year={2022}
}

@misc{deepseekai2025deepseekv32,
      title={DeepSeek-V3.2: Pushing the Frontier of Open Large Language Models}, 
      author={DeepSeek-AI},
      year={2025},
}

@misc{5team2025glm45agenticreasoningcoding,
      title={GLM-4.5: Agentic, Reasoning, and Coding (ARC) Foundation Models}, 
      author={GLM Team and Aohan Zeng and Xin Lv and Qinkai Zheng and Zhenyu Hou and Bin Chen and Chengxing Xie and Cunxiang Wang and Da Yin and Hao Zeng and Jiajie Zhang and Kedong Wang and Lucen Zhong and Mingdao Liu and Rui Lu and Shulin Cao and Xiaohan Zhang and Xuancheng Huang and Yao Wei and Yean Cheng and Yifan An and Yilin Niu and Yuanhao Wen and Yushi Bai and Zhengxiao Du and Zihan Wang and Zilin Zhu and Bohan Zhang and Bosi Wen and Bowen Wu and Bowen Xu and Can Huang and Casey Zhao and Changpeng Cai and Chao Yu and Chen Li and Chendi Ge and Chenghua Huang and Chenhui Zhang and Chenxi Xu and Chenzheng Zhu and Chuang Li and Congfeng Yin and Daoyan Lin and Dayong Yang and Dazhi Jiang and Ding Ai and Erle Zhu and Fei Wang and Gengzheng Pan and Guo Wang and Hailong Sun and Haitao Li and Haiyang Li and Haiyi Hu and Hanyu Zhang and Hao Peng and Hao Tai and Haoke Zhang and Haoran Wang and Haoyu Yang and He Liu and He Zhao and Hongwei Liu and Hongxi Yan and Huan Liu and Huilong Chen and Ji Li and Jiajing Zhao and Jiamin Ren and Jian Jiao and Jiani Zhao and Jianyang Yan and Jiaqi Wang and Jiayi Gui and Jiayue Zhao and Jie Liu and Jijie Li and Jing Li and Jing Lu and Jingsen Wang and Jingwei Yuan and Jingxuan Li and Jingzhao Du and Jinhua Du and Jinxin Liu and Junkai Zhi and Junli Gao and Ke Wang and Lekang Yang and Liang Xu and Lin Fan and Lindong Wu and Lintao Ding and Lu Wang and Man Zhang and Minghao Li and Minghuan Xu and Mingming Zhao and Mingshu Zhai and Pengfan Du and Qian Dong and Shangde Lei and Shangqing Tu and Shangtong Yang and Shaoyou Lu and Shijie Li and Shuang Li and Shuang-Li and Shuxun Yang and Sibo Yi and Tianshu Yu and Wei Tian and Weihan Wang and Wenbo Yu and Weng Lam Tam and Wenjie Liang and Wentao Liu and Xiao Wang and Xiaohan Jia and Xiaotao Gu and Xiaoying Ling and Xin Wang and Xing Fan and Xingru Pan and Xinyuan Zhang and Xinze Zhang and Xiuqing Fu and Xunkai Zhang and Yabo Xu and Yandong Wu and Yida Lu and Yidong Wang and Yilin Zhou and Yiming Pan and Ying Zhang and Yingli Wang and Yingru Li and Yinpei Su and Yipeng Geng and Yitong Zhu and Yongkun Yang and Yuhang Li and Yuhao Wu and Yujiang Li and Yunan Liu and Yunqing Wang and Yuntao Li and Yuxuan Zhang and Zezhen Liu and Zhen Yang and Zhengda Zhou and Zhongpei Qiao and Zhuoer Feng and Zhuorui Liu and Zichen Zhang and Zihan Wang and Zijun Yao and Zikang Wang and Ziqiang Liu and Ziwei Chai and Zixuan Li and Zuodong Zhao and Wenguang Chen and Jidong Zhai and Bin Xu and Minlie Huang and Hongning Wang and Juanzi Li and Yuxiao Dong and Jie Tang},
      year={2025},
      eprint={2508.06471},
      archivePrefix={arXiv},
      primaryClass={cs.CL},
      url={https://arxiv.org/abs/2508.06471}, 
}

@inproceedings{kwon2023efficient,
  title={Efficient Memory Management for Large Language Model Serving with PagedAttention},
  author={Woosuk Kwon and Zhuohan Li and Siyuan Zhuang and Ying Sheng and Lianmin Zheng and Cody Hao Yu and Joseph E. Gonzalez and Hao Zhang and Ion Stoica},
  booktitle={Proceedings of the ACM SIGOPS 29th Symposium on Operating Systems Principles},
  year={2023}
}

@inproceedings{tao2025treerag,
  title={Treerag: Unleashing the power of hierarchical storage for enhanced knowledge retrieval in long documents},
  author={Tao, Wenyu and Xing, Xiaofen and Chen, Yirong and Huang, Linyi and Xu, Xiangmin},
  booktitle={Findings of the Association for Computational Linguistics: ACL 2025},
  pages={356--371},
  year={2025}
}

@inproceedings{sarthi2024raptor,
  title={Raptor: Recursive abstractive processing for tree-organized retrieval},
  author={Sarthi, Parth and Abdullah, Salman and Tuli, Aditi and Khanna, Shubh and Goldie, Anna and Manning, Christopher D},
  booktitle={The Twelfth International Conference on Learning Representations},
  year={2024}
}

@article{li2025search,
  title={Search-o1: Agentic search-enhanced large reasoning models},
  author={Li, Xiaoxi and Dong, Guanting and Jin, Jiajie and Zhang, Yuyao and Zhou, Yujia and Zhu, Yutao and Zhang, Peitian and Dou, Zhicheng},
  journal={arXiv preprint arXiv:2501.05366},
  year={2025}
}

@article{cui2025paddleocr,
  title={Paddleocr-vl: Boosting multilingual document parsing via a 0.9 b ultra-compact vision-language model},
  author={Cui, Cheng and Sun, Ting and Liang, Suyin and Gao, Tingquan and Zhang, Zelun and Liu, Jiaxuan and Wang, Xueqing and Zhou, Changda and Liu, Hongen and Lin, Manhui and others},
  journal={arXiv preprint arXiv:2510.14528},
  year={2025}
}

@article{wei2025deepseek,
  title={Deepseek-ocr: Contexts optical compression},
  author={Wei, Haoran and Sun, Yaofeng and Li, Yukun},
  journal={arXiv preprint arXiv:2510.18234},
  year={2025}
}

@article{zhang2025pageindex,
  author = {Mingtian Zhang and Yu Tang and PageIndex Team},
  title = {PageIndex: Next-Generation Vectorless, Reasoning-based RAG},
  journal = {PageIndex Blog},
  year = {2025},
  month = {September},
  note = {https://pageindex.ai/blog/pageindex-intro},
}

@article{jin2025search,
  title={Search-r1: Training llms to reason and leverage search engines with reinforcement learning},
  author={Jin, Bowen and Zeng, Hansi and Yue, Zhenrui and Yoon, Jinsung and Arik, Sercan and Wang, Dong and Zamani, Hamed and Han, Jiawei},
  journal={arXiv preprint arXiv:2503.09516},
  year={2025}
}

@article{team2025hunyuanocr,
  title={HunyuanOCR Technical Report},
  author={Team, Hunyuan Vision and Lyu, Pengyuan and Wan, Xingyu and Li, Gengluo and Peng, Shangpin and Wang, Weinong and Wu, Liang and Shen, Huawen and Zhou, Yu and Tang, Canhui and others},
  journal={arXiv preprint arXiv:2511.19575},
  year={2025}
}

@article{wang2024mineru,
  title={Mineru: An open-source solution for precise document content extraction},
  author={Wang, Bin and Xu, Chao and Zhao, Xiaomeng and Ouyang, Linke and Wu, Fan and Zhao, Zhiyuan and Xu, Rui and Liu, Kaiwen and Qu, Yuan and Shang, Fukai and others},
  journal={arXiv preprint arXiv:2409.18839},
  year={2024}
}

@article{li2025dots,
  title={dots. ocr: Multilingual document layout parsing in a single vision-language model},
  author={Li, Yumeng and Yang, Guang and Liu, Hao and Wang, Bowen and Zhang, Colin},
  journal={arXiv preprint arXiv:2512.02498},
  year={2025}
}


\clearpage
\appendix

\section{Appendix}
\subsection{Benchmark Statistics}
Table~\ref{tab:dataset_statistics} details the statistics of the four datasets employed in our evaluation. The benchmarks are categorized into single-document and multi-document settings to assess the agent's performance across different retrieval scopes. Notably, the single-document datasets pose a significant challenge regarding context length: FinanceBench averages approximately 165k tokens, while our constructed ContextBench reaches an average of 233k tokens, serving as a rigorous stress test for long-document reasoning capabilities.
\begin{table}[h]
    \centering
    \caption{Statistics of the Datasets Used in Evaluation. The token counts are calculated based on the parsed Markdown content.}
    \label{tab:dataset_statistics}
    \resizebox{\columnwidth}{!}{%
        \begin{tabular}{llrrr}
            \toprule
            \textbf{Dataset} & \textbf{Type} & \textbf{\# Examples} & \textbf{Total Tokens} & \textbf{Avg. Tokens} \\
            \midrule
            FinanceBench~\cite{islam2023financebench} & Single-Doc & 150 & 24,725,584 & 164,837 \\
            ContextBench (Ours) & Single-Doc & 94 & 21,671,771 & 233,030 \\
            \midrule
            QASPER~\cite{dasigi2021dataset} & Multi-Doc & 143 & 6,437,147 & 45,015 \\
            SyllabusQA~\cite{fernandez2024syllabusqa} & Multi-Doc & 196 & 2,800,293 & 14,287 \\
            \bottomrule
        \end{tabular}%
    }
\end{table}
\subsection{Benchmark Construction Details}
\label{ap:bench_detail}
In this paper, we manually annotated the ContextBench long-document QA dataset for real-world scenarios. This annotation effort was inspired by the challenge that traditional RAG systems face when segmenting documents into chunks. During this process, certain contiguous contexts are split into two or more chunks, diluting the original semantic relationships during representation. This poses significant challenges for retrieval systems. Specifically, we recruited 12 researchers with expertise in natural language processing and a solid understanding of LLMs and RAG. We first provided annotators with several golden examples—comprising documents, questions, and answers—along with the evidence distribution within the documents. This allowed annotators to grasp the challenges involved. Subsequently, they selected questions requiring long-context and long-range dependencies from long documents they encountered in their daily work. Each of the 12 annotators labeled 10 questions. After manual review, we obtained 94 samples. On average, each sample required 0.5 person-hours of effort. 

Regarding our approach to synthesizing multi-document QA datasets, specifically, following the prompt sequence in Figure~\ref{fig:qa_gen_prompt}, we instructed the LLM to generate questions based on the full context of multiple documents. We then manually reviewed each question to ensure quality. Ultimately, QASPER yielded 143 samples, and SyllabusQA produced 196 samples.
\subsection{Case Study}
\label{app:Case}
We conduct fine-grained case analysis on four benchmarks to illustrate the core value of the \texttt{ReadSection} tool in DeepRead's locate-then-read paradigm:

\textbf{ContextBench (Table~\ref{tab:case_cb}).}
For the query ``Which Agents are there in the Analyst Team?'', DeepRead first locates the relevant section (\texttt{doc\_id=1, sec\_id=13}) via semantic retrieval, then invokes \texttt{ReadSection} to read paragraphs 0--8 of this section contiguously. This operation retrieves the complete, structured list of four Analyst Team agents (Fundamental/Sentiment/News/Technical Analyst Agents) and their responsibilities---information that would be fragmented or incomplete if relying solely on sparse retrieval snippets. The continuous reading capability of \texttt{ReadSection} ensures the stable extraction of structured lists (e.g., team member composition), a key advantage over structure-blind retrieval baselines.

\textbf{FinanceBench (Table~\ref{tab:case_fb}).}
When calculating Amazon's FY2016--2017 revenue growth, semantic retrieval initially returns non consolidated income statement data (2016: \$152,283M, 2017: \$187,890M), leading to an incorrect 23.4\% growth rate. The critical 7th round \texttt{ReadSection} call accesses the consolidated statements of operations, retrieving the true net sales figures (2016: \$135,987M, 2017: \$177,866M) and enabling the correct 30.8\% calculation. This demonstrates \texttt{ReadSection}'s role in validating and correcting fragmented retrieval results by accessing complete, authoritative document sections.

\textbf{QASPER (Table~\ref{tab:case_qasper}).}
For multi-document queries about dataset specifications (e.g., TIMIT acoustic embeddings, BookTest construction), structure-blind retrieval often misses granular details (e.g., TIMIT's 630 speakers, 80-dimensional Mel filter banks; BookTest's Gutenberg source and cloze generation rules). DeepRead's 5th and 9th round \texttt{ReadSection} calls fill these gaps by reading contiguous sections of relevant documents, ensuring accurate, complete answers instead of guesswork or omission of key dataset attributes.

\textbf{SyllabusQA (Table~\ref{tab:case_syllabus}).}
For the query about late work policies in course syllabi, semantic retrieval only returns grading tables (e.g., exam weights) but misses critical late work rules. The 5th round \texttt{ReadSection} call accesses the full course policy section, extracting rules like ``Late work is not accepted except in special circumstances''---information that is non-tabular, context-dependent, and unretrievable via keyword-based snippet matching. This highlights \texttt{ReadSection}'s value in capturing unstructured but semantically critical policy details.

Tables~\ref{tab:case_cb}--\ref{tab:case_qasper} present complete interaction trajectories for these cases, confirming that \texttt{ReadSection} complements retrieval by reconstructing contiguous, context-rich evidence. This enables DeepRead to mimic human reading patterns: localizing key sections via lightweight retrieval, then deep-reading to consolidate complete, accurate information---a paradigm that outperforms structure-blind retrieval baselines in capturing both structured lists and unstructured policy details.

\subsection{Robustness Testing of LLM as a Judge}
\label{app:judge}
\newcommand{\J}[3]{\texttt{#1}/\texttt{#2}/\texttt{#3}}
\newcommand{\AG}[1]{\texttt{#1}}

\begin{table*}[t]
\centering
\caption{\textbf{Accuracy (\%) under three independent LLM judges.}
Each entry is reported as \textbf{DeepSeek-V3.2 / GLM-4.7 / Qwen3-235B} (in this order).
}
\label{tab:judge_replication}
\scriptsize
\resizebox{\textwidth}{!}{
\begin{tabular}{lccccccc}
\toprule
\multirow{2}{*}{\textbf{Setting}} & \multicolumn{3}{c}{\textbf{Single-Document}} & \multicolumn{3}{c}{\textbf{Multi-Document}} & \multirow{2}{*}{\textbf{Overall Avg}} \\
\cmidrule(lr){2-4} \cmidrule(lr){5-7}
& \textbf{FinanceBench} & \textbf{ContextBench} & \textbf{Avg} & \textbf{QASPER} & \textbf{SyllabusQA} & \textbf{Avg} & \\
\midrule
Search-o1 & \J{80.0}{80.7}{84.0} & \J{74.5}{71.3}{70.2} & \J{77.2}{76.0}{77.1} & \J{65.0}{58.7}{55.2} & \J{57.1}{55.6}{58.7} & \J{61.1}{57.2}{57.0} & \J{69.2}{66.6}{67.0} \\
Search-o1 w/ expand & \J{83.3}{83.3}{85.3} & \J{84.0}{79.8}{80.9} & \J{83.7}{81.6}{83.1} & \J{65.0}{56.6}{60.1} & \J{68.4}{67.3}{65.8} & \J{66.7}{62.0}{63.0} & \J{75.2}{71.8}{73.0} \\
\hline
DeepRead & \J{82.7}{84.0}{88.7} & \J{91.5}{87.2}{91.5} & \J{87.1}{85.6}{90.1} & \J{72.7}{69.9}{65.7} & \J{70.9}{71.4}{71.4} & \J{71.8}{70.7}{68.6} & \J{79.5}{78.1}{79.3} \\
DeepRead w/ expand & \J{84.0}{84.0}{88.0} & \J{88.3}{84.0}{87.2} & \J{86.1}{84.0}{87.6} & \J{76.2}{63.6}{62.2} & \J{72.4}{69.4}{73.0} & \J{74.3}{66.5}{67.6} & \J{80.3}{75.3}{77.6} \\
\hline
Read-only & \J{80.7}{84.0}{89.3} & \J{91.5}{87.2}{86.2} & \J{86.1}{85.6}{87.8} & \J{18.9}{12.6}{15.4} & \J{15.3}{18.9}{22.4} & \J{17.1}{15.7}{18.9} & \J{51.6}{50.7}{53.3} \\
\bottomrule
\end{tabular}
}
\end{table*}

\begin{table*}[t]
\centering
\caption{\textbf{Inter-judge agreement (higher is more consistent).}
Agreement is computed by our evaluation script and reflects example-level consistency of the three judges' binary verdicts (correct/incorrect).}
\label{tab:judge_agreement}
\scriptsize
\resizebox{\textwidth}{!}{
\begin{tabular}{lccccccc}
\toprule
\multirow{2}{*}{\textbf{Setting}} & \multicolumn{3}{c}{\textbf{Single-Document}} & \multicolumn{3}{c}{\textbf{Multi-Document}} & \multirow{2}{*}{\textbf{Overall Avg}} \\
\cmidrule(lr){2-4} \cmidrule(lr){5-7}
& \textbf{FinanceBench} & \textbf{ContextBench} & \textbf{Avg} & \textbf{QASPER} & \textbf{SyllabusQA} & \textbf{Avg} & \\
\midrule
Search-o1 & \AG{0.9400} & \AG{0.9043} & \AG{0.9221} & \AG{0.8462} & \AG{0.8622} & \AG{0.8542} & \AG{0.8882} \\
Search-o1 w/ expand & \AG{0.9200} & \AG{0.9043} & \AG{0.9121} & \AG{0.8182} & \AG{0.8214} & \AG{0.8198} & \AG{0.8660} \\
\hline
DeepRead & \AG{0.9133} & \AG{0.9255} & \AG{0.9194} & \AG{0.8671} & \AG{0.8827} & \AG{0.8749} & \AG{0.8972} \\
DeepRead w/ expand & \AG{0.9267} & \AG{0.9362} & \AG{0.9314} & \AG{0.7902} & \AG{0.8724} & \AG{0.8313} & \AG{0.8814} \\
\hline
Read-only & \AG{0.8800} & \AG{0.9362} & \AG{0.9081} & \AG{0.8601} & \AG{0.9082} & \AG{0.8842} & \AG{0.8961} \\
\midrule
\textbf{Avg (over settings)} & \AG{0.9160} & \AG{0.9213} & \AG{0.9186} & \AG{0.8364} & \AG{0.8694} & \AG{0.8529} & \AG{0.8858} \\
\bottomrule
\end{tabular}
}
\end{table*}
To ensure the robustness of our conclusions, we replicate the evaluation using three independent LLMs: DeepSeek-V3.2, GLM-4.7, and Qwen3-235B-A22B-thinking-2507. The mean accuracies across all settings are highly consistent (70.53\%, 67.90\%, and 69.59\% respectively), confirming that the observed improvements are not artifacts of a specific judge's calibration. Furthermore, we compute the inter-judge agreement to validate evaluation reliability. We define the agreement score as the proportion of samples where all three judges reach a unanimous verdict. Let $J_m(x_i)$ be the verdict of judge $m$ on sample $i$. The metric is calculated as:
\begin{equation}
    \text{Agreement} = \frac{1}{N} \sum_{i=1}^{N} \mathbb{I}(J_1(x_i) = J_2(x_i) = J_3(x_i))
\end{equation}
where $\mathbb{I}(\cdot)$ denotes the indicator function. The system achieves a high overall agreement of \textbf{0.8858}. As expected, agreement is higher on single-document tasks (avg. 0.9187) compared to multi-document tasks (avg. 0.8540), reflecting the inherent complexity and slight subjectivity involved in evaluating cross-document reasoning.
\subsection{Prompt Template}
In order to ensure reproducibility and transparency within our methodology, we provide a detailed description of the prompt templates in this paper. Figure~\ref{fig:system_prompt} presents the System Prompt. Figure~\ref{fig:judge_prompt} illustrates the LLM-as-a-Judge prompt, which is designed to assess the correctness of AI-generated answers relative to a human-generated ``golden answer''. Finally, Figure~\ref{fig:qa_gen_prompt} showcases the prompt used for generating multi-hop question-answer pairs for our benchmarks. This prompt is particularly stringent, as it enforces constraints on both hierarchical dependency and cross-document reasoning. The requirement for questions to synthesize information from at least four distinct leaf sections ensures that the generated QA pairs reflect complex reasoning and comprehensive document understanding. Furthermore, the closed-form nature of the questions, requiring a single unambiguous answer, maintains precision in evaluation.

\begin{table*}[t]
    \centering
    \caption{An example from DeepRead on ContextBench. The model-generated semantic retrieval queries are enclosed within \green{<|begin\_semantic\_retrieval\_query|>} and \green{<|end\_semantic\_retrieval\_query|>}, while the top-2 retrieval results are enclosed within \blue{<|begin\_semantic\_retrieval\_result|>} and \blue{<|end\_semantic\_retrieval\_result|>}. Similarly, the read section queries are enclosed within \yellow{<|begin\_read\_section\_query|>} and \yellow{<|end\_read\_section\_query|>}, and the read section results are enclosed within \red{<|begin\_read\_section\_result|>} and \red{<|end\_read\_section\_result|>}.}
    \fontsize{9pt}{11pt}\selectfont
    \begin{tabular}{p{0.98\linewidth}}
    \midrule
        \rowcolor{gray!20}\textbf{Example \#1 from ContextBench } \\
    \midrule
        \textbf{Question:} \\
        Which Agents are there in the Analyst Team? \\
        \textbf{Labeled Answer:} \\
        There are four Agents in the Analyst Team: \\(1) Fundamental Analyst Agents\\(2) Sentiment Analyst Agents\\(3) News Analyst Agents\\(4) Technical Analyst Agents.\\
    \midrule
        \rowcolor{gray!20}
        \textbf{Model Output by DeepRead}\\
    \midrule
        
        \textbf{ROUND 1 (Retrieval)}\\
        I'll help you find information about the agents in the Analyst Team. Let me search for relevant information in the document.\\
        \green{<|begin\_semantic\_retrieval\_query|>}Analyst Team agents members\green{<|end\_semantic\_retrieval\_query|>}\\
        
        \blue{<|begin\_semantic\_retrieval\_result(top1)|>}doc\_id=1, sec\_id=13, para\_id=0, 
        text=``The Analyst Team (Figure 2) is composed of specialized agents responsible for gathering and analyzing various types of market data to inform trading decisions. Each agent focuses on a specific aspect of market analysis, bringing together a comprehensive view of the market’s conditions.''\blue{<|end\_semantic\_retrieval\_result(top1)|>}\\

        \blue{<|begin\_semantic\_retrieval\_result(top2)|>}doc\_id=1, sec\_id=15, para\_id=0, 
        text=``Trader Agents (Figure 4) are responsible for executing trading decisions based on the comprehensive analysis provided by the Analyst Team and the nuanced perspectives from the Researcher Team. They''\blue{<|end\_semantic\_retrieval\_result(top2)|>}\\\\
    
        \textbf{ROUND 2 (ReadSection)}\\
        Now let me read the section about the Analyst Team to get detailed information about the agents it contains.\\
        
        \yellow{<|begin\_read\_section\_query|>}doc\_id=1, sec\_id=13, start\_para\_id=0, end\_para\_id=8\yellow{<|end\_read\_section\_query|>}\\
        
        \red{<|begin\_read\_section\_result(para\_id=0)|>}The Analyst Team (Figure 2) is composed of specialized agents responsible for gathering and analyzing various types of market data to inform trading decisions. ……\red{<|end\_read\_section\_result(para\_id=0)|>}\\
        ……\\
        \red{<|begin\_read\_section\_result(para\_id=3)|>}• Fundamental Analyst Agents: These agents evaluate company fundamentals by analyzing financial statements, earnings reports, insider transactions, and other pertinent data. They assess a company’s intrinsic value to identify undervalued or overvalued stocks, providing insights into long-term investment potential.\red{<|end\_read\_section\_result(para\_id=3)|>}\\
        \red{<|begin\_read\_section\_result(para\_id=4)|>}- Sentiment Analyst Agents: These agents process large volumes of social media posts, sentiment scores, and insider sentiments derived from public information and social media activity. They gauge market sentiment to predict how collective investor behavior might impact stock prices in the short term.\red{<|end\_read\_section\_result(para\_id=4)|>}\\

        \red{<|begin\_read\_section\_result(para\_id=5)|>}• News Analyst Agents: These agents analyze news articles, government announcements, and other macroeconomic indicators to assess the market’s macroeconomic state, major world events, and significant company changes. They identify news events that could influence market movements, helping to anticipate sudden shifts in market dynamics.\red{<|end\_read\_section\_result(para\_id=5)|>}\\
        \red{<|begin\_read\_section\_result(para\_id=6)|>}• Technical Analyst Agents: These agents calculate and select relevant technical indicators, such as Moving Average Convergence Divergence (MACD) and Relative Strength Index (RSI), customized for specific assets. They analyze price patterns and trading volumes to forecast future price movements, assisting in timing entry and exit points.\red{<|end\_read\_section\_result(para\_id=6)|>}

        \red{<|begin\_read\_section\_result(para\_id=7)|>}Collectively, the Analyst Team synthesizes data from multiple sources to provide a holistic market analysis. ……\red{<|end\_read\_section\_result(para\_id=7)|>}\\
        \bottomrule
        \end{tabular}
        \label{tab:case_cb}
\end{table*}

\begin{table*}[t]
    \ContinuedFloat
    \centering
    \caption{An example from DeepRead on ContextBench. The model-generated semantic retrieval queries are enclosed within \green{<|begin\_semantic\_retrieval\_query|>} and \green{<|end\_semantic\_retrieval\_query|>}, while the top-2 retrieval results are enclosed within \blue{<|begin\_semantic\_retrieval\_result|>} and \blue{<|end\_semantic\_retrieval\_result|>}. Similarly, the read section queries are enclosed within \yellow{<|begin\_read\_section\_query|>} and \yellow{<|end\_read\_section\_query|>}, and the read section results are enclosed within \red{<|begin\_read\_section\_result|>} and \red{<|end\_read\_section\_result|>}. (Continued)}
    \fontsize{9pt}{11pt}\selectfont
    \begin{tabular}{p{0.98\linewidth}}
    \midrule
        \rowcolor{gray!20}\textbf{Example \#1 from ContextBench (Continued)} \\
    \midrule
        \textbf{Question:} \\
        Which Agents are there in the Analyst Team? \\
        \textbf{Labeled Answer:} \\
        There are four Agents in the Analyst Team: \\(1) Fundamental Analyst Agents\\(2) Sentiment Analyst Agents\\(3) News Analyst Agents\\(4) Technical Analyst Agents.\\
    \midrule
        \rowcolor{gray!20}
        \textbf{Model Output by DeepRead}\\
    \midrule
    
        \textbf{ROUND 3 (Final Answer)}\\
        Based on the document, the Analyst Team in TradingAgents consists of four specialized agents:\\1. **Fundamental Analyst Agents** - These agents evaluate company fundamentals by analyzing financial statements, earnings reports, insider transactions, and other pertinent data to assess a company's intrinsic value and identify undervalued or overvalued stocks.\\2. **Sentiment Analyst Agents** - These agents process social media posts, sentiment scores, and insider sentiments to gauge market sentiment and predict how collective investor behavior might impact stock prices in the short term.\\3. **News Analyst Agents** - These agents analyze news articles, government announcements, and macroeconomic indicators to assess the market's macroeconomic state, major world events, and significant company changes.\\4. **Technical Analyst Agents** - These agents calculate and select relevant technical indicators (such as MACD and RSI), analyze price patterns and trading volumes to forecast future price movements, and assist in timing entry and exit points.\\The Analyst Team works collectively to synthesize data from multiple sources and provide holistic market analysis, which then serves as foundational input for the Researcher Team in the trading decision-making process.\\
        \bottomrule
        \end{tabular}
        \label{tab:case_cb}
\end{table*}

\begin{table*}[t]
    \centering
    \caption{An example from DeepRead on FinanceBench. The model-generated semantic retrieval queries are enclosed within \green{<|begin\_semantic\_retrieval\_query|>} and \green{<|end\_semantic\_retrieval\_query|>}, while the top-2 retrieval results are enclosed within \blue{<|begin\_semantic\_retrieval\_result|>} and \blue{<|end\_semantic\_retrieval\_result|>}. Similarly, the read section queries are enclosed within \yellow{<|begin\_read\_section\_query|>} and \yellow{<|end\_read\_section\_query|>}, and the read section results are enclosed within \red{<|begin\_read\_section\_result|>} and \red{<|end\_read\_section\_result|>}.}
    \label{tab:case_fb}
    \fontsize{9pt}{11pt}\selectfont
    \begin{tabular}{p{0.98\linewidth}}
    \toprule
        \rowcolor{gray!20}\textbf{Example \#2 from FinanceBench } \\
    \midrule
        \textbf{Question:} \\
        What is Amazon's year-over-year change in revenue from FY2016 to FY2017 (in units of percents and round to one decimal place)? Calculate what was asked by utilizing the line items clearly shown in the statement of income.\\
        \textbf{Labeled Answer:} \\
        30.8\%\\
    \midrule
        \rowcolor{gray!20}
        \textbf{Model Output by DeepRead}\\
    \midrule
        
        \textbf{ROUND 1 (Retrieval)}\\
        I'll help you find Amazon's year-over-year revenue change from FY2016 to FY2017. First, I need to locate the financial statements and specifically the income statement data for those years.\\
        \green{<|begin\_semantic\_retrieval\_query|>}Amazon revenue 2016 2017 year over year change income statement\green{<|end\_semantic\_retrieval\_query|>}\\
        
        \blue{<|begin\_semantic\_retrieval\_result(top1)|>}doc\_id=1, sec\_id=62, para\_id=6, 
        text=``The increase in AWS operating income in absolute dollars in 2016 and 2017, compared to the comparable prior year periods, is primarily due to increased customer usage and cost structure productivity, partially offset by pricing changes and increased spending on technology infrastructure and sales and marketing expenses and related payroll, which was primarily driven by additional investments to support the business growth. Changes in foreign exchange rates impacted operating income by \$264 million, \$(5) million, and \$(53) million for 2015, 2016, and 2017.''\blue{<|end\_semantic\_retrieval\_result(top1)|>}\\

        \blue{<|begin\_semantic\_retrieval\_result(top2)|>}doc\_id=1, sec\_id=62, para\_id=4, 
        text=``The increase in North America operating income in absolute dollars in 2016 and 2017, compared to the comparable prior year periods, is primarily due to increased unit sales, including sales by third-party sellers, partially offset by increased levels of operating expenses to expand our fulfillment network and spending on technology and content and marketing efforts. Changes in foreign exchange rates impacted operating income by \$30 million, \$27 million, and \$(4) million for 2015, 2016, and 2017.''\blue{<|end\_semantic\_retrieval\_result(top2)|>}\\\\

        \textbf{ROUND 2 (Retrieval)}\\
        I need to find the actual revenue numbers from the income statement. Let me search for the consolidated financial statements and specifically look for net sales or revenue figures.\\
        
        \green{<|begin\_semantic\_retrieval\_query|>}consolidated income statement net sales 2016 2017\green{<|end\_semantic\_retrieval\_query|>}\\
        
        \blue{<|begin\_semantic\_retrieval\_result(top1)|>}doc\_id=1, sec\_id=131, para\_id=1, 
        text=``<table border=1 style='margin: auto; width: max-content;'><tr><td rowspan="2"></td><td colspan="3">Twelve Months Ended December 31,</td></tr><tr><td style='text-align: center;'>2016</td><td style='text-align: center;'>2017</td><td style='text-align: center;'></td></tr><tr><td style='text-align: center;'>Net sales</td><td style='text-align: center;'>\$152,283</td><td style='text-align: center;'>\$187,890</td><td style='text-align: center;'></td></tr><tr><td style='text-align: center;'>Net income</td><td style='text-align: center;'>\$2,148</td><td style='text-align: center;'>\$2,940</td><td style='text-align: center;'></td></tr></table>''\blue{<|end\_semantic\_retrieval\_result(top1)|>}\\

        \blue{<|begin\_semantic\_retrieval\_result(top2)|>}doc\_id=1, sec\_id=131, para\_id=0, 
        text=``The acquired companies were consolidated into our financial statements starting on their respective acquisition dates. The aggregate net sales and operating loss of Whole Foods Market consolidated into our financial statements since the date of acquisition was \$5.8 billion and \$(24) million for the year ended December 31, 2017. The aggregate net sales and operating loss of other acquisitions consolidated into our financial statements since the respective dates of acquisition was \$482 million and \$(142) million for the year ended December 31, 2017. The following financial information, which excludes certain acquired companies for which the pro forma impact is not meaningful, presents our results as if the acquisitions during the year ended December 31, 2017 had occurred on January 1, 2016 (in millions):''\blue{<|end\_semantic\_retrieval\_result(top2)|>}\\
    
    \bottomrule
    \end{tabular}
\end{table*}

\begin{table*}[t]
    \ContinuedFloat
    \centering
    \caption{An example from DeepRead on FinanceBench. The model-generated semantic retrieval queries are enclosed within \green{<|begin\_semantic\_retrieval\_query|>} and \green{<|end\_semantic\_retrieval\_query|>}, while the top-2 retrieval results are enclosed within \blue{<|begin\_semantic\_retrieval\_result|>} and \blue{<|end\_semantic\_retrieval\_result|>}. Similarly, the read section queries are enclosed within \yellow{<|begin\_read\_section\_query|>} and \yellow{<|end\_read\_section\_query|>}, and the read section results are enclosed within \red{<|begin\_read\_section\_result|>} and \red{<|end\_read\_section\_result|>}. (Continued)}
    \label{tab:case_fb}
    \fontsize{9pt}{11pt}\selectfont
    \begin{tabular}{p{0.98\linewidth}}
    \toprule
        \rowcolor{gray!20}\textbf{Example \#2 from FinanceBench (Continued) } \\
    \midrule
        \textbf{Question:} \\
        What is Amazon's year-over-year change in revenue from FY2016 to FY2017 (in units of percents and round to one decimal place)? Calculate what was asked by utilizing the line items clearly shown in the statement of income.\\
        \textbf{Labeled Answer:} \\
        30.8\%\\
    \midrule
        \rowcolor{gray!20}
        \textbf{Model Output by DeepRead}\\
    \midrule
        
        \textbf{ROUND 3 (Retrieval)}\\
        Great! I found the net sales figures for 2016 and 2017. According to the table in node 131:\\- 2016 Net sales: \$152,283 million\\- 2017 Net sales: \$187,890 million\\\\Now I need to calculate the year-over-year change in percentage terms and round to one decimal place.\\\\Let me calculate:\\1. Calculate the difference: \$187,890 - \$152,283 = \$35,607 million\\2. Calculate the percentage change: (\$35,607 / \$152,283) × 100 = 23.38\%\\\\Rounded to one decimal place: 23.4\%\\\\However, I should verify these numbers are from the actual consolidated income statement, not just pro forma data. Let me search for the actual consolidated income statement.\\
        \green{<|begin\_semantic\_retrieval\_query|>}consolidated statements of operations net sales 2016 2017\green{<|end\_semantic\_retrieval\_query|>}\\
        
        \blue{<|begin\_semantic\_retrieval\_result(top1)|>}doc\_id=1, sec\_id=131, para\_id=1, 
        text=``<table border=1 style='margin: auto; width: max-content;'><tr><td rowspan="2"></td><td colspan="3">Twelve Months Ended December 31,</td></tr><tr><td style='text-align: center;'>2016</td><td style='text-align: center;'>2017</td><td style='text-align: center;'></td></tr><tr><td style='text-align: center;'>Net sales</td><td style='text-align: center;'>\$152,283</td><td style='text-align: center;'>\$187,890</td><td style='text-align: center;'></td></tr><tr><td style='text-align: center;'>Net income</td><td style='text-align: center;'>\$2,148</td><td style='text-align: center;'>\$2,940</td><td style='text-align: center;'></td></tr></table>''\blue{<|end\_semantic\_retrieval\_result(top1)|>}\\

        \blue{<|begin\_semantic\_retrieval\_result(top2)|>}doc\_id=1, sec\_id=131, para\_id=0, 
        text=``The acquired companies were consolidated into our financial statements starting on their respective acquisition dates. The aggregate net sales and operating loss of Whole Foods Market consolidated into our financial statements since the date of acquisition was \$5.8 billion and \$(24) million for the year ended December 31, 2017. The aggregate net sales and operating loss of other acquisitions consolidated into our financial statements since the respective dates of acquisition was \$482 million and \$(142) million for the year ended December 31, 2017. The following financial information, which excludes certain acquired companies for which the pro forma impact is not meaningful, presents our results as if the acquisitions during the year ended December 31, 2017 had occurred on January 1, 2016 (in millions):''\blue{<|end\_semantic\_retrieval\_result(top2)|>}\\
    \bottomrule
    \end{tabular}
\end{table*}

\begin{table*}[t]
    \ContinuedFloat
    \centering
    \caption{An example from DeepRead on FinanceBench. The model-generated semantic retrieval queries are enclosed within \green{<|begin\_semantic\_retrieval\_query|>} and \green{<|end\_semantic\_retrieval\_query|>}, while the top-2 retrieval results are enclosed within \blue{<|begin\_semantic\_retrieval\_result|>} and \blue{<|end\_semantic\_retrieval\_result|>}. Similarly, the read section queries are enclosed within \yellow{<|begin\_read\_section\_query|>} and \yellow{<|end\_read\_section\_query|>}, and the read section results are enclosed within \red{<|begin\_read\_section\_result|>} and \red{<|end\_read\_section\_result|>}. (Continued)}
    \label{tab:case_fb}
    \fontsize{9pt}{11pt}\selectfont
    \begin{tabular}{p{0.98\linewidth}}
    \toprule
        \rowcolor{gray!20}\textbf{Example \#2 from FinanceBench (Continued) } \\
    \midrule
        \textbf{Question:} \\
        What is Amazon's year-over-year change in revenue from FY2016 to FY2017 (in units of percents and round to one decimal place)? Calculate what was asked by utilizing the line items clearly shown in the statement of income.\\
        \textbf{Labeled Answer:} \\
        30.8\%\\
    \midrule
        \rowcolor{gray!20}
        \textbf{Model Output by DeepRead}\\
    \midrule
        \textbf{ROUND 4 (ReadSection)}\\
        I need to find the actual consolidated income statement, not pro forma data. Let me look for the actual financial statements. Based on the directory structure, I should look at the consolidated balance sheets or income statements. Let me check node 90 which appears to be the consolidated balance sheets.\\
        
        \yellow{<|begin\_read\_section\_query|>}doc\_id=1, sec\_id=90, start\_para\_id=0, end\_para\_id=5\yellow{<|end\_read\_section\_query|>}\\
        
        \red{<|begin\_read\_section\_result(para\_id=0)|>}<table border=1 style='margin: auto; width: max-content;'><tr><td rowspan="2"></td><td colspan="3">December 31,</td></tr><tr><td style='text-align: center;'>2016</td><td style='text-align: center;'>2017</td><td style='text-align: center;'></td></tr><tr><td style='text-align: center;'>ASSETS</td><td style='text-align: center;'></td><td style='text-align: center;'></td><td style='text-align: center;'></td></tr><tr><td style='text-align: center;'>Current assets:</td><td style='text-align: center;'></td><td style='text-align: center;'></td><td style='text-align: center;'></td></tr><tr><td style='text-align: center;'>Cash and cash equivalents</td><td style='text-align: center;'>\$19,334</td>……\red{<|end\_read\_section\_result(para\_id=0)|>}\\
        ……\\
        \red{<|begin\_read\_section\_result(para\_id=4)|>}See accompanying notes to consolidated financial statements.\red{<|end\_read\_section\_result(para\_id=4)|>}\\\\

        \textbf{ROUND5 (Retrieval)}\\
        This is the balance sheet and stockholders' equity statement, not the income statement. I need to find the actual consolidated statements of operations. Let me search more specifically.\\

        \green{<|begin\_semantic\_retrieval\_query|>}consolidated statements of operations 2016 2017 net sales revenue\green{<|end\_semantic\_retrieval\_query|>}\\
        
        \blue{<|begin\_semantic\_retrieval\_result(top1)|>}doc\_id=1, sec\_id=131, para\_id=1, 
        text=``<table border=1 style='margin: auto; width: max-content;'><tr><td rowspan="2"></td><td colspan="3">Twelve Months Ended December 31,</td></tr><tr><td style='text-align: center;'>2016</td><td style='text-align: center;'>2017</td><td style='text-align: center;'></td></tr><tr><td style='text-align: center;'>Net sales</td><td style='text-align: center;'>\$152,283</td><td style='text-align: center;'>\$187,890</td><td style='text-align: center;'></td></tr><tr><td style='text-align: center;'>Net income</td><td style='text-align: center;'>\$2,148</td><td style='text-align: center;'>\$2,940</td><td style='text-align: center;'></td></tr></table>''\blue{<|end\_semantic\_retrieval\_result(top1)|>}\\

        \blue{<|begin\_semantic\_retrieval\_result(top2)|>}doc\_id=1, sec\_id=131, para\_id=0, 
        text=``The acquired companies were consolidated into our financial statements starting on their respective acquisition dates. The aggregate net sales and operating loss of Whole Foods Market consolidated into our financial statements since the date of acquisition was \$5.8 billion and \$(24) million for the year ended December 31, 2017. The aggregate net sales and operating loss of other acquisitions consolidated into our financial statements since the respective dates of acquisition was \$482 million and \$(142) million for the year ended December 31, 2017. The following financial information, which excludes certain acquired companies for which the pro forma impact is not meaningful, presents our results as if the acquisitions during the year ended December 31, 2017 had occurred on January 1, 2016 (in millions):''\blue{<|end\_semantic\_retrieval\_result(top2)|>}\\
    \bottomrule
    \end{tabular}
\end{table*}

\begin{table*}[t]
    \ContinuedFloat
    \centering
    \caption{An example from DeepRead on FinanceBench. The model-generated semantic retrieval queries are enclosed within \green{<|begin\_semantic\_retrieval\_query|>} and \green{<|end\_semantic\_retrieval\_query|>}, while the top-2 retrieval results are enclosed within \blue{<|begin\_semantic\_retrieval\_result|>} and \blue{<|end\_semantic\_retrieval\_result|>}. Similarly, the read section queries are enclosed within \yellow{<|begin\_read\_section\_query|>} and \yellow{<|end\_read\_section\_query|>}, and the read section results are enclosed within \red{<|begin\_read\_section\_result|>} and \red{<|end\_read\_section\_result|>}. (Continued)}
    \label{tab:case_fb}
    \fontsize{9pt}{11pt}\selectfont
    \begin{tabular}{p{0.98\linewidth}}
    \toprule
        \rowcolor{gray!20}\textbf{Example \#2 from FinanceBench (Continued) } \\
    \midrule
        \textbf{Question:} \\
        What is Amazon's year-over-year change in revenue from FY2016 to FY2017 (in units of percents and round to one decimal place)? Calculate what was asked by utilizing the line items clearly shown in the statement of income.\\
        \textbf{Labeled Answer:} \\
        30.8\%\\
    \midrule
        \rowcolor{gray!20}
        \textbf{Model Output by DeepRead}\\
    \midrule
        \textbf{ROUND 6 (ReadSection)}\\
        I need to find the actual consolidated income statement, not pro forma data. Let me look for the actual financial statements. Based on the directory structure, I should look at the consolidated balance sheets or income statements. Let me check node 90 which appears to be the consolidated balance sheets.\\
        
        \yellow{<|begin\_read\_section\_query|>}doc\_id=1, sec\_id=85, start\_para\_id=0, end\_para\_id=1\yellow{<|end\_read\_section\_query|>}\\
        
        \red{<|begin\_read\_section\_result(para\_id=0)|>}<table border=1 style='margin: auto; width: max-content;'><tr><td colspan="2">Page</td>……\red{<|end\_read\_section\_result(para\_id=0)|>}\\\\

        \textbf{ROUND 7 (ReadSection)}\\
        Good! The Consolidated Statements of Operations are on page 38. Now I need to find where that content is in the document structure. Looking at the directory, I see node 89 is "AMAZON.COM, INC" and node 90 is the consolidated balance sheets. The statements of operations should be before the balance sheets. Let me check node 89.\\

        \yellow{<|begin\_read\_section\_query|>}doc\_id=1, sec\_id=89, start\_para\_id=0, end\_para\_id=6\yellow{<|end\_read\_section\_query|>}\\
        
        \red{<|begin\_read\_section\_result(para\_id=0)|>}<div style="text-align: center;">CONSOLIDATED STATEMENTS OF OPERATIONS   (in millions, except per share data)</div>\red{<|end\_read\_section\_result(para\_id=0)|>}\\
        \red{<|begin\_read\_section\_result(para\_id=1)|>}<table border=1 style='margin: auto; width: max-content;'><tr><td rowspan=\"2\"></td><td colspan=\"4\">Year Ended December 31,</td></tr><tr><td style='text-align: center;'>2015</td><td style='text-align: center;'>2016</td><td style='text-align: center;'>2017</td><td style='text-align: center;'></td></tr><tr><td style='text-align: center;'>Net product sales</td><td style='text-align: center;'>\$79,268</td><td style='text-align: center;'>\$94,665</td><td style='text-align: center;'>\$ 118,573</td><td style='text-align: center;'></td></tr><tr><td style='text-align: center;'>Net service sales</td><td style='text-align: center;'>27,738</td><td style='text-align: center;'>41,322</td><td style='text-align: center;'>59,293</td><td style='text-align: center;'></td></tr><tr><td style='text-align: center;'>Total net sales</td><td style='text-align: center;'>107,006</td><td style='text-align: center;'>135,987</td><td style='text-align: center;'>177,866</td>……\red{<|end\_read\_section\_result(para\_id=1)|>}
        ……\\
        \red{<|begin\_read\_section\_result(para\_id=5)|>}See accompanying notes to consolidated financial statements.\red{<|end\_read\_section\_result(para\_id=5)|>}\\\\

        \textbf{ROUND 8 (Final Answer)}\\
        Perfect! I found the actual Consolidated Statements of Operations. From the table, I can see:\\\\- 2016 Total net sales: \$135,987 million\\- 2017 Total net sales: \$177,866 million\\\\Now I need to calculate the year-over-year change in percentage terms and round to one decimal place.\\\\Calculation:\\1. Difference: \$177,866 - \$135,987 = \$41,879 million\\2. Percentage change: (\$41,879 / \$135,987) × 100 = 30.80\%\\\\Rounded to one decimal place: 30.8\%\\ **Answer:** Amazon's year-over-year change in revenue from FY2016 to FY2017 was 30.8\%.\\
    \bottomrule
    \end{tabular}
\end{table*}

\begin{table*}[t]
    \centering
    \caption{An example from DeepRead on SyllabusQA. The model-generated semantic retrieval queries are enclosed within \green{<|begin\_semantic\_retrieval\_query|>} and \green{<|end\_semantic\_retrieval\_query|>}, while the top-2 retrieval results are enclosed within \blue{<|begin\_semantic\_retrieval\_result|>} and \blue{<|end\_semantic\_retrieval\_result|>}. Similarly, the read section queries are enclosed within \yellow{<|begin\_read\_section\_query|>} and \yellow{<|end\_read\_section\_query|>}, and the read section results are enclosed within \red{<|begin\_read\_section\_result|>} and \red{<|end\_read\_section\_result|>}.}
    \fontsize{9pt}{11pt}\selectfont
    \begin{tabular}{p{0.98\linewidth}}
    \midrule
        \rowcolor{gray!20}\textbf{Example \#3 from SyllabusQA } \\
    \midrule
        \textbf{Question:} \\
        What constitutes the total percentage weight of all examinations toward the final grade in the courses that include them?\\
        \textbf{Labeled Answer:} \\
        The total examination weight is 40\% in the Social Entrepreneurship course and 45\% in the Individual Income Taxes course.\\
    \midrule
        \rowcolor{gray!20}
        \textbf{Model Output by DeepRead}\\
    \midrule
        
        \textbf{ROUND 1 (Retrieval)}\\
       I'll help you find information about the examination weight toward the final grade. Let me start by searching for relevant information about grading and exams.\\
        \green{<|begin\_semantic\_retrieval\_query|>}exams weight percentage final grade\green{<|end\_semantic\_retrieval\_query|>}\\
        
        \blue{<|begin\_semantic\_retrieval\_result(top1)|>}doc\_id=4, sec\_id=5, para\_id=1, 
        text=``<table border=1 style='margin: auto; width: max-content;'><tr><td style='text-align: center;'>Weighting</td><td style='text-align: center;'>\%</td><td colspan="2">Grade Cutoffs</td></tr><tr><td style='text-align: center;'>Exam 1</td><td style='text-align: center;'>15\%</td><td style='text-align: center;'>A</td><td style='text-align: center;'>94-100\%</td></tr><tr><td style='text-align: center;'>Exam 2</td><td style='text-align: center;'>15\%</td><td style='text-align: center;'>A-</td><td style='text-align: center;'>90-93\%</td></tr><tr><td style='text-align: center;'>Exam 3</td><td style='text-align: center;'>15\%</td><td style='text-align: center;'>B+</td><td style='text-align: center;'>86-89\%</td></tr><tr><td style='text-align: center;'>Tax Return Problem</td><td style='text-align: center;'>15\%</td><td style='text-align: center;'>B</td><td style='text-align: center;'>82-85\%</td></tr><tr><td style='text-align: center;'>Pre-Class Quiz</td><td style='text-align: center;'>10\%</td><td style='text-align: center;'>B-</td><td style='text-align: center;'>79-81\%</td></tr><tr><td style='text-align: center;'>Homework</td><td style='text-align: center;'>10\%</td><td style='text-align: center;'>C+</td><td style='text-align: center;'>75-78\%</td></tr><tr><td style='text-align: center;'>Quizzes</td><td style='text-align: center;'>10\%</td><td style='text-align: center;'>C</td><td style='text-align: center;'>72-74\%</td></tr><tr><td style='text-align: center;'>In-Class Assignments</td><td style='text-align: center;'>10\%</td><td style='text-align: center;'>C-</td><td style='text-align: center;'>69-71\%</td></tr><tr><td style='text-align: center;'></td><td style='text-align: center;'>—</td><td style='text-align: center;'>D+</td><td style='text-align: center;'>66-68\%</td></tr><tr><td style='text-align: center;'>Total</td><td style='text-align: center;'>100\%</td><td style='text-align: center;'>D</td><td style='text-align: center;'>60-65\%</td></tr><tr><td style='text-align: center;'></td><td style='text-align: center;'></td><td style='text-align: center;'>F</td><td style='text-align: center;'>Below 60\%</td></tr></table>''\blue{<|end\_semantic\_retrieval\_result(top1)|>}\\

        \blue{<|begin\_semantic\_retrieval\_result(top2)|>}doc\_id=2, sec\_id=9, para\_id=5, 
        text=``<table border=1 style='margin: auto; width: max-content;'><tr><td style='text-align: center;'>Assignment</td><td style='text-align: center;'>Percentage of Final Grade</td></tr><tr><td style='text-align: center;'>Analytics assignments (4)</td><td style='text-align: center;'>80\%</td></tr><tr><td style='text-align: center;'>Online Reflections (8)</td><td style='text-align: center;'>20\%</td></tr><tr><td style='text-align: center;'></td><td style='text-align: center;'>100\%</td></tr></table>''\blue{<|end\_semantic\_retrieval\_result(top2)|>}\\
        \bottomrule
        \end{tabular}
        \label{tab:case_syllabus}
\end{table*}

\begin{table*}[t]
    \ContinuedFloat
    \centering
    \caption{An example from DeepRead on SyllabusQA. The model-generated semantic retrieval queries are enclosed within \green{<|begin\_semantic\_retrieval\_query|>} and \green{<|end\_semantic\_retrieval\_query|>}, while the top-2 retrieval results are enclosed within \blue{<|begin\_semantic\_retrieval\_result|>} and \blue{<|end\_semantic\_retrieval\_result|>}. Similarly, the read section queries are enclosed within \yellow{<|begin\_read\_section\_query|>} and \yellow{<|end\_read\_section\_query|>}, and the read section results are enclosed within \red{<|begin\_read\_section\_result|>} and \red{<|end\_read\_section\_result|>}. (Continued)}
    \fontsize{9pt}{11pt}\selectfont
    \begin{tabular}{p{0.98\linewidth}}
    \midrule
        \rowcolor{gray!20}\textbf{Example \#3 from SyllabusQA (Continued)} \\
    \midrule
        \textbf{Question:} \\
        What constitutes the total percentage weight of all examinations toward the final grade in the courses that include them?\\
        \textbf{Labeled Answer:} \\
        The total examination weight is 40\% in the Social Entrepreneurship course and 45\% in the Individual Income Taxes course.\\
    \midrule
        \rowcolor{gray!20}
        \textbf{Model Output by DeepRead}\\
    \midrule
        \textbf{ROUND 2 (ReadSection)}\\
        I can see there's a grading table in document 4, node 5. Let me read that section to get the complete information about exam weights.\\
        
        \yellow{<|begin\_read\_section\_query|>}doc\_id=4, sec\_id=5, start\_para\_id=0, end\_para\_id=2\yellow{<|end\_read\_section\_query|>}\\
        
        \red{<|begin\_read\_section\_result(para\_id=0)|>}Your grade will be based on a number of factors, which will be weighted and assigned according to the following chart:\red{<|end\_read\_section\_result(para\_id=0)|>}\\
        
        \red{<|begin\_read\_section\_result(para\_id=1)|>}<table border=1 style='margin: auto; width: max-content;'><tr><td style='text-align: center;'>Weighting</td><td style='text-align: center;'>\%</td><td colspan="2">Grade Cutoffs</td></tr><tr><td style='text-align: center;'>Exam 1</td><td style='text-align: center;'>15\%</td><td style='text-align: center;'>A</td><td style='text-align: center;'>94-100\%</td></tr><tr><td style='text-align: center;'>Exam 2</td><td style='text-align: center;'>15\%</td><td style='text-align: center;'>A-</td><td style='text-align: center;'>90-93\%</td></tr><tr><td style='text-align: center;'>Exam 3</td><td style='text-align: center;'>15\%</td><td style='text-align: center;'>B+</td><td style='text-align: center;'>86-89\%</td></tr><tr><td style='text-align: center;'>Tax Return Problem</td><td style='text-align: center;'>15\%</td><td style='text-align: center;'>B</td><td style='text-align: center;'>82-85\%</td></tr><tr><td style='text-align: center;'>Pre-Class Quiz</td><td style='text-align: center;'>10\%</td><td style='text-align: center;'>B-</td><td style='text-align: center;'>79-81\%</td></tr><tr><td style='text-align: center;'>Homework</td><td style='text-align: center;'>10\%</td><td style='text-align: center;'>C+</td><td style='text-align: center;'>75-78\%</td></tr><tr><td style='text-align: center;'>Quizzes</td><td style='text-align: center;'>10\%</td><td style='text-align: center;'>C</td><td style='text-align: center;'>72-74\%</td></tr><tr><td style='text-align: center;'>In-Class Assignments</td><td style='text-align: center;'>10\%</td><td style='text-align: center;'>C-</td><td style='text-align: center;'>69-71\%</td></tr><tr><td style='text-align: center;'></td><td style='text-align: center;'>—</td><td style='text-align: center;'>D+</td><td style='text-align: center;'>66-68\%</td></tr><tr><td style='text-align: center;'>Total</td><td style='text-align: center;'>100\%</td><td style='text-align: center;'>D</td><td style='text-align: center;'>60-65\%</td></tr><tr><td style='text-align: center;'></td><td style='text-align: center;'></td><td style='text-align: center;'>F</td><td style='text-align: center;'>Below 60\%</td></tr></table>\red{<|end\_read\_section\_result(para\_id=1)|>}\\
        \bottomrule
        \end{tabular}
        \label{tab:case_syllabus}
\end{table*}

\begin{table*}[t]
    \ContinuedFloat
    \centering
    \caption{An example from DeepRead on SyllabusQA. The model-generated semantic retrieval queries are enclosed within \green{<|begin\_semantic\_retrieval\_query|>} and \green{<|end\_semantic\_retrieval\_query|>}, while the top-2 retrieval results are enclosed within \blue{<|begin\_semantic\_retrieval\_result|>} and \blue{<|end\_semantic\_retrieval\_result|>}. Similarly, the read section queries are enclosed within \yellow{<|begin\_read\_section\_query|>} and \yellow{<|end\_read\_section\_query|>}, and the read section results are enclosed within \red{<|begin\_read\_section\_result|>} and \red{<|end\_read\_section\_result|>}. (Continued)}
    \fontsize{9pt}{11pt}\selectfont
    \begin{tabular}{p{0.98\linewidth}}
    \midrule
        \rowcolor{gray!20}\textbf{Example \#3 from SyllabusQA (Continued)} \\
    \midrule
        \textbf{Question:} \\
        What constitutes the total percentage weight of all examinations toward the final grade in the courses that include them?\\
        \textbf{Labeled Answer:} \\
        The total examination weight is 40\% in the Social Entrepreneurship course and 45\% in the Individual Income Taxes course.\\
    \midrule
        \rowcolor{gray!20}
        \textbf{Model Output by DeepRead}\\
    \midrule
        \textbf{ROUND 3 (Retrieval)}\\
       Now let me check if there are any other courses or documents that mention exams to get a complete picture. Let me search for other exam-related information.\\
        \green{<|begin\_semantic\_retrieval\_query|>}examination exam final grade weight\green{<|end\_semantic\_retrieval\_query|>}\\
        
        \blue{<|begin\_semantic\_retrieval\_result(top1)|>}doc\_id=4, sec\_id=5, para\_id=1, 
        text=``<table border=1 style='margin: auto; width: max-content;'><tr><td style='text-align: center;'>Weighting</td><td style='text-align: center;'>\%</td><td colspan="2">Grade Cutoffs</td></tr><tr><td style='text-align: center;'>Exam 1</td><td style='text-align: center;'>15\%</td><td style='text-align: center;'>A</td><td style='text-align: center;'>94-100\%</td></tr><tr><td style='text-align: center;'>Exam 2</td><td style='text-align: center;'>15\%</td><td style='text-align: center;'>A-</td><td style='text-align: center;'>90-93\%</td></tr><tr><td style='text-align: center;'>Exam 3</td><td style='text-align: center;'>15\%</td><td style='text-align: center;'>B+</td><td style='text-align: center;'>86-89\%</td></tr><tr><td style='text-align: center;'>Tax Return Problem</td><td style='text-align: center;'>15\%</td><td style='text-align: center;'>B</td><td style='text-align: center;'>82-85\%</td></tr><tr><td style='text-align: center;'>Pre-Class Quiz</td><td style='text-align: center;'>10\%</td><td style='text-align: center;'>B-</td><td style='text-align: center;'>79-81\%</td></tr><tr><td style='text-align: center;'>Homework</td><td style='text-align: center;'>10\%</td><td style='text-align: center;'>C+</td><td style='text-align: center;'>75-78\%</td></tr><tr><td style='text-align: center;'>Quizzes</td><td style='text-align: center;'>10\%</td><td style='text-align: center;'>C</td><td style='text-align: center;'>72-74\%</td></tr><tr><td style='text-align: center;'>In-Class Assignments</td><td style='text-align: center;'>10\%</td><td style='text-align: center;'>C-</td><td style='text-align: center;'>69-71\%</td></tr><tr><td style='text-align: center;'></td><td style='text-align: center;'>—</td><td style='text-align: center;'>D+</td><td style='text-align: center;'>66-68\%</td></tr><tr><td style='text-align: center;'>Total</td><td style='text-align: center;'>100\%</td><td style='text-align: center;'>D</td><td style='text-align: center;'>60-65\%</td></tr><tr><td style='text-align: center;'></td><td style='text-align: center;'></td><td style='text-align: center;'>F</td><td style='text-align: center;'>Below 60\%</td></tr></table>''\blue{<|end\_semantic\_retrieval\_result(top1)|>}\\

        \blue{<|begin\_semantic\_retrieval\_result(top2)|>}doc\_id=2, sec\_id=9, para\_id=5, 
        text=``<table border=1 style='margin: auto; width: max-content;'><tr><td style='text-align: center;'>Assignment</td><td style='text-align: center;'>Percentage of Final Grade</td></tr><tr><td style='text-align: center;'>Analytics assignments (4)</td><td style='text-align: center;'>80\%</td></tr><tr><td style='text-align: center;'>Online Reflections (8)</td><td style='text-align: center;'>20\%</td></tr><tr><td style='text-align: center;'></td><td style='text-align: center;'>100\%</td></tr></table>''\blue{<|end\_semantic\_retrieval\_result(top2)|>}\\\\
    
        \textbf{ROUND 4 (ReadSection)}\\
        Let me also check the "Exams" section in document 4 to see if there's more information about exams.\\
        
        \yellow{<|begin\_read\_section\_query|>}doc\_id=4, sec\_id=6, start\_para\_id=0, end\_para\_id=1\yellow{<|end\_read\_section\_query|>}\\
        
        \red{<|begin\_read\_section\_result(para\_id=0)|>}The exams will be outside of class time to reduce time stress. I will describe the layout of the exams beforehand. The exams will be open book. If you miss an exam I will give a make-up exam for extreme reasons, be prepared to take a make-up as soon as possible. I follow the rules of the Academic Regulations issued by the University. These rules require written proof of an acceptable absence. With that said, I will need written proof if you are not able to make an exam. I cannot give makeups simply because you feel inadequately prepared as that seems unfair to the class as a whole. Please contact me in advance if you will miss an exam. The exams will be in the evenings and the times will be confirmed as soon as possible. I will follow the University final exam schedule so please do not schedule yourself to be unavailable during finals week.\red{<|end\_read\_section\_result(para\_id=0)|>}\\
        \bottomrule
        \end{tabular}
        \label{tab:case_syllabus}
\end{table*}

\begin{table*}[h]
    \ContinuedFloat
    \centering
    \caption{An example from DeepRead on SyllabusQA. The model-generated semantic retrieval queries are enclosed within \green{<|begin\_semantic\_retrieval\_query|>} and \green{<|end\_semantic\_retrieval\_query|>}, while the top-2 retrieval results are enclosed within \blue{<|begin\_semantic\_retrieval\_result|>} and \blue{<|end\_semantic\_retrieval\_result|>}. Similarly, the read section queries are enclosed within \yellow{<|begin\_read\_section\_query|>} and \yellow{<|end\_read\_section\_query|>}, and the read section results are enclosed within \red{<|begin\_read\_section\_result|>} and \red{<|end\_read\_section\_result|>}. (Continued)}
    \fontsize{9pt}{11pt}\selectfont
    \begin{tabular}{p{0.98\linewidth}}
    \midrule
        \rowcolor{gray!20}\textbf{Example \#3 from SyllabusQA (Continued)} \\
    \midrule
        \textbf{Question:} \\
        What constitutes the total percentage weight of all examinations toward the final grade in the courses that include them?\\
        \textbf{Labeled Answer:} \\
        The total examination weight is 40\% in the Social Entrepreneurship course and 45\% in the Individual Income Taxes course.\\
    \midrule
        \rowcolor{gray!20}
        \textbf{Model Output by DeepRead}\\
    \midrule
        \textbf{ROUND 5 (ReadSection)}\\
       Now let me check document 2 to see if it contains any information about exams or grading. Let me read the relevant section from document 2.\\
        \yellow{<|begin\_read\_section\_query|>}doc\_id=2, sec\_id=9, start\_para\_id=0, end\_para\_id=8\yellow{<|end\_read\_section\_query|>}\\
        
        \red{<|begin\_read\_section\_result(para\_id=0)|>}Each module has a designated discussion forum for questions about the module's contents. Students are expected to post their questions about the course contents to the designated module discussion forum. ……\red{<|end\_read\_section\_result(para\_id=0)|>}\\
        ……\\
        \red{<|begin\_read\_section\_result(para\_id=2)|>}In order to receive full credit for work, students must turn in required deliverables on the specified due date. Late work is not accepted except in special circumstances (e.g., religious holidays, military duty, jury duty, etc.) or other extremely unusual circumstances. Extremely unusual circumstances must be discussed with the instructor PRIOR to due dates. Incomplete grades are not given. Please plan accordingly and finish all work during the scope of this course.\red{<|end\_read\_section\_result(para\_id=2)|>}
        ……\\
        \red{<|begin\_read\_section\_result(para\_id=7)|>}<table border=1 style='margin: auto; width: max-content;'><tr><td style='text-align: center;'>Percent</td><td style='text-align: center;'>Grade</td>……\red{<|end\_read\_section\_result(para\_id=7)|>}\\\\

        \textbf{ROUND 6 (Final Answer)}\\
        Based on the information I've gathered from the documents, I can now answer your question about the total percentage weight of all examinations toward the final grade.\\\\From document 4 (ACCOUNTING 371-01/02 - Individual Income Taxes), the grading table shows:\\- Exam 1: 15\%\\- Exam 2: 15\% \\- Exam 3: 15\%\\\\The total weight of all examinations (Exam 1 + Exam 2 + Exam 3) is 15\% + 15\% + 15\% = **45\%**.\\\\From document 2 (Academic Resources), there is no mention of traditional examinations. The grading is based on:\\- Analytics assignments (4): 80\%\\- Online Reflections (8): 20\%\\\\This course does not include traditional examinations, so the examination weight is **0\%**.\\\\Therefore, in the courses that include examinations (specifically ACCOUNTING 371), the total percentage weight of all examinations toward the final grade is **45\%**.\\
        \bottomrule
        \end{tabular}
        \label{tab:case_syllabus}
\end{table*}

\begin{table*}[t]
    \centering
    \caption{An example from DeepRead on QASPER. The model-generated semantic retrieval queries are enclosed within \green{<|begin\_semantic\_retrieval\_query|>} and \green{<|end\_semantic\_retrieval\_query|>}, while the top-2 retrieval results are enclosed within \blue{<|begin\_semantic\_retrieval\_result|>} and \blue{<|end\_semantic\_retrieval\_result|>}. Similarly, the read section queries are enclosed within \yellow{<|begin\_read\_section\_query|>} and \yellow{<|end\_read\_section\_query|>}, and the read section results are enclosed within \red{<|begin\_read\_section\_result|>} and \red{<|end\_read\_section\_result|>}.}
    \fontsize{9pt}{11pt}\selectfont
    \begin{tabular}{p{0.98\linewidth}}
    \midrule
        \rowcolor{gray!20}\textbf{Example \#4 from QASPER } \\
    \midrule
        \textbf{Question:} \\
        According to the experimental results, what are the reported accuracy rates for Support Vector Machines (SVM) in Vietnamese word segmentation, and how does the perplexity of the standard Transformer baseline compare to the proposed Global Variational Transformer (GVT) on the MojiTalk dataset?\\
        \textbf{Labeled Answer:} \\
        SVMs achieved accuracy rates of 97.2\% and 94.05\%. On the MojiTalk dataset, the standard Transformer had a perplexity of 72.66, which is higher than the GVT's perplexity of 19.71.\\
    \midrule
        \rowcolor{gray!20}
        \textbf{Model Output by DeepRead}\\
    \midrule
        
        \textbf{ROUND 1 (Retrieval)}\\
       I'll help you find the information about SVM accuracy rates in Vietnamese word segmentation and the perplexity comparison between Transformer baseline and GVT on the MojiTalk dataset. Let me start by searching for relevant information.\\
       
        \green{<|begin\_semantic\_retrieval\_query|>}Support Vector Machines SVM accuracy Vietnamese word segmentation\green{<|end\_semantic\_retrieval\_query|>}\\
        
        \blue{<|begin\_semantic\_retrieval\_result(top1)|>}doc\_id=2, sec\_id=2, para\_id=5, 
        text=``There are several studies about the Vietnamese word segmentation task over the last decade. Dinh et al. started this task with Weighted Finite State Transducer (WFST) approach and Neural Network approach$[10]$ . In addition, machine learning approaches are studied and widely applied to natural language processing and word segmentation as well. In fact, several studies used support vector machines (SVM) and conditional random fields (CRF) for the word segmentation task $[8]$ , $[9]$ . Based on annotated corpora and token-based features, studies used machine learning approaches to build word segmentation systems with accuracy about 94\%-97\%.''\blue{<|end\_semantic\_retrieval\_result(top1)|>}\\

        \blue{<|begin\_semantic\_retrieval\_result(top2)|>}doc\_id=2, sec\_id=13, para\_id=2, 
        text=``This technique is applied for Vietnamese word segmentation in several studies $[8]$ , $[25]$ . Nguyen et al. applied on a segmented corpus of 8,000 sentences and got the result at 94.05\% while Ngo et al. used it with 45,531 segmented sentences and got the result at 97.2\%. It is worth to mention that general SVM libraries (such as LIBSVM$^{4}$ , LIBLINEAR$^{5}$ , SVMlight$^{6}$ , Node-SVM$^{7}$ , and TreeSVM$^{8}$ ), YamCha$^{9}$ is an opened source SVM library that serves several NLP tasks: POS tagging, Named Entity Recognition, base NP chunking, Text Chunking, Text Classification and event Word Segmentation.''\blue{<|end\_semantic\_retrieval\_result(top2)|>}\\\\
    
        \textbf{ROUND 2 (Retrieval)}\\
        Now let me search for information about perplexity comparison between Transformer baseline and GVT on the MojiTalk dataset.\\

        \green{<|begin\_semantic\_retrieval\_query|>}Transformer baseline perplexity Global Variational Transformer GVT MojiTalk dataset\green{<|end\_semantic\_retrieval\_query|>}\\
        
        \blue{<|begin\_semantic\_retrieval\_result(top1)|>}doc\_id=1, sec\_id=20, para\_id=3, 
        text=``<table border=1 style='margin: auto; width: max-content;'><tr><td colspan="10">MojiTalk</td></tr><tr><td rowspan="2">Model</td>……''\blue{<|end\_semantic\_retrieval\_result(top1)|>}\\

        \blue{<|begin\_semantic\_retrieval\_result(top2)|>}doc\_id=1, sec\_id=17, para\_id=1, 
        text=``MojiTalk dataset consists of 596,959 post and response pairs from Twitter. Each response is labeled by one emoji which indicates the response emotion. There are 64 emoji labels in total with unbalanced distribution. We use the preprocessed data and vocabulary released from Zhou and Wang (2018) and follow the same split of train/validation/test set.''\blue{<|end\_semantic\_retrieval\_result(top2)|>}\\

        \bottomrule
        \end{tabular}
        \label{tab:case_qasper}
\end{table*}

\begin{table*}[!t]
    \ContinuedFloat
    \centering
    \caption[]{An example from DeepRead on QASPER. The model-generated semantic retrieval queries are enclosed within \green{<|begin\_semantic\_retrieval\_query|>} and \green{<|end\_semantic\_retrieval\_query|>}, while the top-2 retrieval results are enclosed within \blue{<|begin\_semantic\_retrieval\_result|>} and \blue{<|end\_semantic\_retrieval\_result|>}. Similarly, the read section queries are enclosed within \yellow{<|begin\_read\_section\_query|>} and \yellow{<|end\_read\_section\_query|>}, and the read section results are enclosed within \red{<|begin\_read\_section\_result|>} and \red{<|end\_read\_section\_result|>}. (Continued)}
    \fontsize{9pt}{11pt}\selectfont
    \begin{tabular}{p{0.98\linewidth}}
    \toprule
        \rowcolor{gray!20}\textbf{Example \#4 from QASPER (Continued)} \\
    \midrule
        \textbf{Question:} \\
        According to the experimental results, what are the reported accuracy rates for Support Vector Machines (SVM) in Vietnamese word segmentation, and how does the perplexity of the standard Transformer baseline compare to the proposed Global Variational Transformer (GVT) on the MojiTalk dataset?\\
        \textbf{Labeled Answer:} \\
        SVMs achieved accuracy rates of 97.2\% and 94.05\%. On the MojiTalk dataset, the standard Transformer had a perplexity of 72.66, which is higher than the GVT's perplexity of 19.71.\\
    \midrule
        \rowcolor{gray!20}
        \textbf{Model Output by DeepRead}\\
    \midrule
        \textbf{ROUND 3 (ReadSection)}\\
        Now let me read the specific section about SVM accuracy rates to get the exact numbers mentioned in the search results.\\
        \yellow{<|begin\_read\_section\_query|>}doc\_id=2, sec\_id=13, start\_para\_id=0, end\_para\_id=3\yellow{<|end\_read\_section\_query|>}\\
        
        \red{<|begin\_read\_section\_result(para\_id=0)|>}Support Vector Machines (SVM) is a supervised machine learning method which considers dataset as a set of vectors and tries to classify them into specific classes. Basically, SVM is a binary classifier. However, most classification tasks are multi-class classifiers. When applying SVMs, the method has been extended to classify three or more classes. Particular NLP tasks, like word segmentation and Part-of-speech task, each token/word in documents will be used as a feature vector. For\red{<|end\_read\_section\_result(para\_id=0)|>}\\
        ……\\
        \red{<|begin\_read\_section\_result(para\_id=2)|>}This technique is applied for Vietnamese word segmentation in several studies $[8]$ , $[25]$ . Nguyen et al. applied on a segmented corpus of 8,000 sentences and got the result at 94.05\% while Ngo et al. used it with 45,531 segmented sentences and got the result at 97.2\%. It is worth to mention that general SVM libraries (such as LIBSVM$^{4}$ , LIBLINEAR$^{5}$ , SVMlight$^{6}$ , Node-SVM$^{7}$ , and TreeSVM$^{8}$ ), YamCha$^{9}$ is an opened source SVM library that serves several NLP tasks: POS tagging, Named Entity Recognition, base NP chunking, Text Chunking, Text Classification and event Word Segmentation.\red{<|end\_read\_section\_result(para\_id=2)|>}\\\\
        
        \textbf{ROUND 4 (ReadSection)}\\
        Now let me also check the introduction section to see if there are more general accuracy ranges mentioned for SVM.\\
        \yellow{<|begin\_read\_section\_query|>}doc\_id=2, sec\_id=2, start\_para\_id=4, end\_para\_id=6\yellow{<|end\_read\_section\_query|>}\\
        
        \red{<|begin\_read\_section\_result(para\_id=4)|>}Image\red{<|end\_read\_section\_result(para\_id=4)|>}\\
        \red{<|begin\_read\_section\_result(para\_id=5)|>}There are several studies about the Vietnamese word segmentation task over the last decade. Dinh et al. started this task with Weighted Finite State Transducer (WFST) approach and Neural Network approach $[10]$ . In addition, machine learning approaches are studied and widely applied to natural language processing and word segmentation as well. In fact, several studies used support vector machines (SVM) and conditional random fields (CRF) for the word segmentation task $[8]$ , $[9]$. Based on annotated corpora and token-based features, studies used machine learning approaches to build word segmentation systems with accuracy about 94\%-97\%.\red{<|end\_read\_section\_result(para\_id=5)|>}\\
        
    \bottomrule
    \end{tabular}
\end{table*}

\begin{table*}[!t]
    \ContinuedFloat
    \centering
    \caption[]{An example from DeepRead on QASPER. The model-generated semantic retrieval queries are enclosed within \green{<|begin\_semantic\_retrieval\_query|>} and \green{<|end\_semantic\_retrieval\_query|>}, while the top-2 retrieval results are enclosed within \blue{<|begin\_semantic\_retrieval\_result|>} and \blue{<|end\_semantic\_retrieval\_result|>}. Similarly, the read section queries are enclosed within \yellow{<|begin\_read\_section\_query|>} and \yellow{<|end\_read\_section\_query|>}, and the read section results are enclosed within \red{<|begin\_read\_section\_result|>} and \red{<|end\_read\_section\_result|>}. (Continued)}
    \fontsize{9pt}{11pt}\selectfont
    \begin{tabular}{p{0.98\linewidth}}
    \toprule
        \rowcolor{gray!20}\textbf{Example \#4 from QASPER (Continued)} \\
    \midrule
        \textbf{Question:} \\
        According to the experimental results, what are the reported accuracy rates for Support Vector Machines (SVM) in Vietnamese word segmentation, and how does the perplexity of the standard Transformer baseline compare to the proposed Global Variational Transformer (GVT) on the MojiTalk dataset?\\
        \textbf{Labeled Answer:} \\
        SVMs achieved accuracy rates of 97.2\% and 94.05\%. On the MojiTalk dataset, the standard Transformer had a perplexity of 72.66, which is higher than the GVT's perplexity of 19.71.\\
    \midrule
        \rowcolor{gray!20}
        \textbf{Model Output by DeepRead}\\
    \midrule
        
        \textbf{ROUND 5 (Final Answer)}\\
        Now I have the information needed to answer both parts of the question. Based on the search results:\\\\1. **SVM accuracy rates for Vietnamese word segmentation**: \\   - The document states that studies using machine learning approaches (including SVM) achieved accuracy rates of about 94\%-97\%.\\   - Specifically, two studies are mentioned: Nguyen et al. achieved 94.05\% accuracy on a corpus of 8,000 sentences, and Ngo et al. achieved 97.2\% accuracy on a corpus of 45,531 sentences.\\\\2. **Perplexity comparison on MojiTalk dataset**:\\   - From the table in the search results, the standard Transformer baseline has a perplexity (PPL) of 72.66.\\   - The proposed Global Variational Transformer (GVT) has a perplexity of 19.71.\\   - This shows that GVT significantly outperforms the Transformer baseline, with GVT achieving much lower perplexity (19.71 vs 72.66).\\\\**Answer**:\\According to the experimental results:\\1. For Support Vector Machines (SVM) in Vietnamese word segmentation, the reported accuracy rates are:\\   - General range: About 94\%-97\%\\   - Specific studies: Nguyen et al. achieved 94.05\% accuracy on 8,000 sentences, and Ngo et al. achieved 97.2\% accuracy on 45,531 sentences.\\\\2. On the MojiTalk dataset:\\   - The standard Transformer baseline has a perplexity of 72.66\\   - The proposed Global Variational Transformer (GVT) has a perplexity of 19.71\\   - GVT significantly outperforms the Transformer baseline with much lower perplexity (approximately 3.7 times better).\\
        
    \bottomrule
    \end{tabular}
\end{table*}
\clearpage
\begin{figure*}[b]
    \centering
    \begin{promptbox}[System Prompt Template]
        You are a documents assistant and will receive one or more documents structured as follows:
        \newline
        \texttt{- (doc\_id) [node\_id] Title | paragraphs=Num | tokens=Num | children=[ID list]}.
        \newline
        Use this structure and your available tools to answer the user's question.
        
        \textbf{\#\# Guidelines}
        \begin{itemize}
            \setlength\itemsep{0em} 
            \item Use \textit{<Search/Retrieve>} to locate relevant nodes based on the directory.
            \item Answer strictly based on the provided corpus; do not fabricate.
            \item The hierarchical structure of documents is represented in the Directory Structure. Parsing errors may cause body text to be mistakenly treated as hierarchical elements (or headings), rendering the heading text inaccessible to search and reading tools. Please make reasonable inferences based on the structure and the content returned by the tool.
            \item Respond in the User's language; align queries with the Directory Structure.
            \item Usually, you need to think step by step and then call tools to locate or read, iterating in this way until you can answer the question.
            \item When calling tools, DO NOT write tool invocations in plain text. Use the structured tool call interface (tool\_calls) only.
        \end{itemize}
        
        \textbf{\#\# Directory Structure}
        \newline
        \textit{<Serialized Table of Contents inserted here...>}
    \end{promptbox}
    \caption{The system prompt used in DeepRead. It injects the hierarchical document skeleton (Directory Structure).}
    \label{fig:system_prompt}
\end{figure*}
\begin{figure*}[b]
    \centering
    \begin{promptbox}[LLM-as-a-Judge Prompt Template]
        You are an expert evaluator for AI-generated responses to queries. Your task is to determine whether the AI-generated answer correctly answers the query based on the golden answer provided by a human expert.
        
        \textbf{Numerical Accuracy:}
        \begin{itemize}
            \setlength\itemsep{0em}
            \item Rounding differences should be \textbf{ignored} if they do not meaningfully change the conclusion.
            \item You can allow some flexibility in accuracy. For example, 1.2 is considered similar to 1.23. Two numbers are considered similar if one can be rounded to the other.
            \item Fractions, percentage, and numerics could be considered similar, for example: "11 of 14" is considered equivalent to "79\%" and "0.79".
        \end{itemize}
        
        \textbf{Evaluation Criteria:}
        \begin{itemize}
            \setlength\itemsep{0em}
            \item If the golden answer or any of its equivalence can be inferred or generated from the AI-generated answer, then the AI-generated answer is considered correct.
            \item If any number, percentage, fraction, or figure in the golden answer is not present in the AI-generated answer, but can be inferred or generated from the AI-generated answer or implicitly exist in the AI-generated answer, then the AI-generated answer is considered correct.
            \item The AI-generated answer is considered correct if it conveys the same or similar meaning, conclusion, or rationale as the golden answer.
            \item If the AI-generated answer is a superset of the golden answer, it is also considered correct.
            \item If the AI-generated answer provides a valid answer or reasonable interpretation compared to the golden answer, it is considered correct.
            \item If the AI-generated answer contains subjective judgments or opinions, it is considered correct as long as they are reasonable and justifiable compared to the golden answer.
            \item Otherwise, the AI-generated answer is incorrect.
        \end{itemize}
        
        \textbf{Inputs:}
        Query: \textit{<question>}
        \begin{itemize}
            \item  AI-Generated Answer: \textit{<predicted\_answer>}
            \item Golden Answer: \textit{<standard\_answer>}
        \end{itemize}
        Your output should be ONLY a boolean value: \texttt{True} or \texttt{False}, nothing else.
    \end{promptbox}
    \caption{The evaluation prompt used for the LLM-as-a-Judge metric. It instructs the evaluator model to focus on semantic equivalence and allow for flexible numerical matching.}
    \label{fig:judge_prompt}
\end{figure*}
\begin{figure*}[b]
    \centering
    \begin{promptbox}[QA Generation Prompt Template]
        \textbf{System Instruction:}
        \newline
        You are a careful question writer tasked with generating hard but answerable CLOSED-FORM QA pairs from MULTIPLE Markdown documents provided in the user message. You MUST use ONLY the provided document TEXT; do not invent facts, do not use images, filenames, URLs, or any external knowledge. Respect the document content.
        
        \textbf{CRITICAL CONSTRAINTS (apply to EACH item):}
        \begin{itemize}
            \setlength\itemsep{0em}
            \item The question must require synthesizing evidence from \textbf{AT LEAST FOUR distinct LEAF sections} (headings with no deeper subheadings). Treat '\#', '\#\#', etc. as hierarchical.
            \item The answer must uniquely follow by integrating information drawn from \textbf{MULTIPLE hierarchy levels} (e.g., a leaf and its ancestors, or leaves under different parents). It must NOT degenerate to fewer than 4 hops.
            \item \textbf{CLOSED-FORM:} The question has a single, unambiguous, concise correct answer present in the text.
            \item The QUESTION must \textbf{NOT} explicitly mention any section titles or heading levels. Phrase naturally (no "as stated in Section 3.2").
            \item Your question must be answered by considering \textbf{all} the provided documents together.
        \end{itemize}
        
        \textbf{OUTPUT FORMAT:}
        \begin{itemize}
            \setlength\itemsep{0em}
            \item Return ONLY valid JSON with exact schema: \texttt{\{"qas": [\{"q": "...", "a": "..."\}, ...]\}}
            \item Do NOT include any other keys. Do NOT wrap in markdown.
        \end{itemize}

        \textbf{User Input Template:}
        \newline
        You will be given \textit{<k>} Markdown documents (The following documents are unordered.) concatenated below.
        \newline
        Generate exactly \textit{<n>} QA pairs that each satisfy the constraints.
        
        DOCUMENTS BEGIN
        \newline
        \textit{<docs\_text>}
        \newline
        DOCUMENTS END
    \end{promptbox}
    \caption{The prompt used to synthesize multi-hop QA pairs.}
    \label{fig:qa_gen_prompt}
\end{figure*}
\end{document}